\documentclass[11pt]{article}

\usepackage[final]{acl}

\usepackage{times}
\usepackage{latexsym}

\usepackage[T1]{fontenc}
\usepackage[utf8]{inputenc}

\usepackage{microtype}

\usepackage{inconsolata}

\usepackage{graphicx}
\usepackage{hyperref}
\usepackage{url}
\usepackage{booktabs}
\usepackage{float}
\usepackage{subcaption}
\usepackage{xcolor}
\usepackage[most]{tcolorbox} % 引入 tcolorbox
\usepackage{multirow}
\usepackage{wrapfig}
\usepackage{array}
\usepackage{placeins} % 提供 \FloatBarrier
\usepackage{cleveref}
\definecolor{skyblue}{RGB}{135, 206, 235}
\definecolor{deeppurple}{RGB}{106, 90, 205}
\usepackage{pifont}
\usepackage{fontawesome5}
\usepackage{stfloats}
\usepackage{enumitem}
\newcommand{\hflogo}{\includegraphics[height=1.2em]{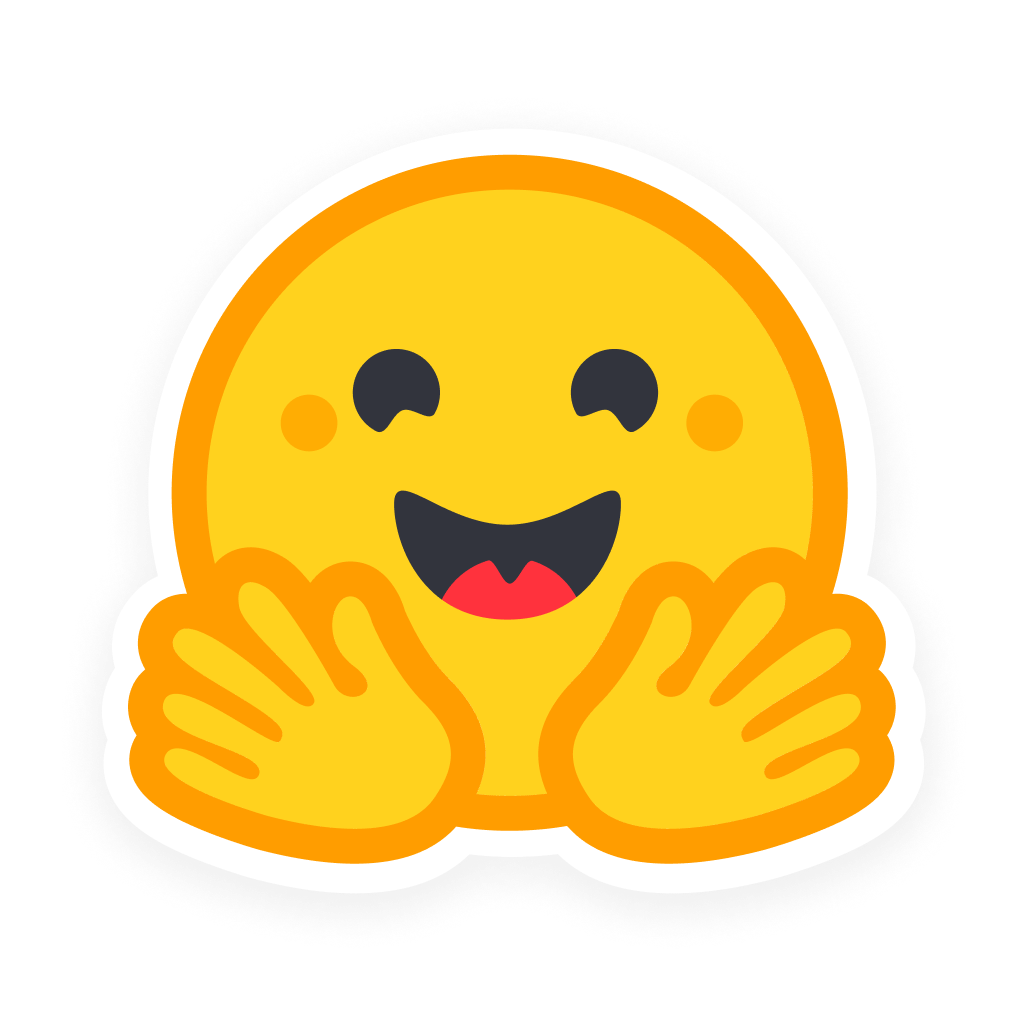}} % Adjust filename

\newcommand{\tzl}[1]{\textcolor{black}{#1}}
\usepackage{amsmath,amsfonts,bm}

\def\eqref#1{equation~\ref{#1}}
\def\1{\bm{1}}

\DeclareMathAlphabet{\mathsfit}{\encodingdefault}{\sfdefault}{m}{sl}
\SetMathAlphabet{\mathsfit}{bold}{\encodingdefault}{\sfdefault}{bx}{n}

\DeclareMathOperator*{\argmin}{arg\,min}

\title{Scaling Behaviors of LLM Reinforcement Learning Post-Training: An Empirical Study in Mathematical Reasoning}

\author{
  \mdseries
  Zelin Tan$^{1,2}$\thanks{This work was done during his internship at Shanghai Artificial Intelligence Laboratory.} \quad
  Hejia Geng$^{3}$ \quad
  Xiaohang Yu$^{4}$ \quad
  Mulei Zhang$^{10}$ \quad
  Guancheng Wan$^{10}$ \\
  Yifan Zhou$^{5}$ \quad
  Qiang He$^{7}$ \quad
  Xiangyuan Xue$^{2,6}$ \quad
  Heng Zhou$^{1,2}$ \quad
  Yutao Fan$^{2}$ \\
  Zhongzhi Li$^{7}$ \quad
  Zaibin Zhang$^{8,3}$ \quad
  Guibin Zhang$^{9}$ \quad
  Chen Zhang$^{2\dagger}$ \quad
  Zhenfei Yin$^{3\dagger}$ \\
  Philip Torr$^{3}$ \quad
  Lei Bai$^{2}$ \\[6pt]
  $^{1}$University of Science and Technology of China \quad
  $^{2}$Shanghai AI Laboratory \quad
  $^{3}$University of Oxford \\
  $^{4}$Imperial College London \quad
  $^{5}$University of Georgia \quad
  $^{6}$The Chinese University of Hong Kong \\
  $^{7}$Chinese Academy of Sciences \quad
  $^{8}$Dalian University of Technology \quad
  $^{9}$National University of Singapore \\
  $^{10}$Wuhan University \\[4pt]
  {\small $^{\dagger}$Corresponding authors: \nolinkurl{zhangchenzc@mail.ustc.edu.cn}, \nolinkurl{jeremyyin@robots.ox.ac.uk}} \\[6pt]
  \href{https://github.com/tanzelin430/Mathematical-Reasoning-RL-Scaling-Law}{%
    \color{black}\faGithub\,\small\texttt{Code}}
  \quad
  \href{https://huggingface.co/datasets/Artemis0430/GURU-MATH-CL}{%
    \color{black}\hflogo\,\small\texttt{Dataset}}
}

\begin{document}
\maketitle
\begin{abstract}
While scaling laws for large language models (LLMs) during pre-training have been extensively studied, their behavior under reinforcement learning (RL) post-training remains largely unexplored. This paper investigates the scaling behavior of Large Language Model (LLM) reinforcement learning post-training, focusing on mathematical reasoning. Through experiments across the Qwen2.5 series (0.5B to 72B), we characterize how model scale, data, and compute interact. Our analysis yields four key findings: \ding{182} Larger models consistently demonstrate superior compute and data efficiency. \ding{183} The relationship between model performance and training resources follows a \textbf{predictive power-law} across both base and instruction-tuned models. \ding{184} RL learning efficiency exhibits a latent \textbf{saturation trend} with increasing model scale.
\ding {185} In data-constrained regimes, performance is primarily driven by the \textbf{total volume of training data} rather than sample uniqueness. These results offer practical guidelines for scaling reasoning capabilities through reinforcement learning post-training.
\end{abstract}
\section{Introduction}
\label{sec:introduction}
The rapid progress of large language models (LLMs) has made elucidating their scaling laws a matter of central importance. These laws, which capture the intricate relationships between model architecture, parameter size, computational cost, data availability, and downstream performance~\citep{kaplan2020scaling, hoffmann2022training}, are invaluable not only because they illuminate the latent factors governing learning dynamics, but also because they provide actionable guidance on how to distribute scarce computational resources most effectively~\citep{li2025misfitting}. While extensive efforts have clarified scaling behavior, the scaling behavior of reinforcement learning (RL) post-training for LLM reasoning remains underexplored.

During pretraining, \citet{kaplan2020scaling} show that cross-entropy loss follows smooth power-law scaling in model size, dataset size, and training compute, implying that larger models trained for fewer steps are compute-optimal. \citet{hoffmann2022training} refine this by showing that, under fixed compute, scaling parameters and tokens proportionally is optimal, since many large models are undertrained. \tzl{Extending to neural-based RL, \citet{hilton2023scaling} empirically demonstrates that the intrinsic performance of convolutional neural networks (CNNs) optimized via reinforcement learning also scales like power-law with model capacity and environment interaction.}

\begin{figure*}[t] % 或者使用 [t!] / [p]
    \centering
    % --- 子图1a ---
    \begin{subfigure}[b]{0.48\textwidth}
        \centering
        \includegraphics[width=\textwidth]{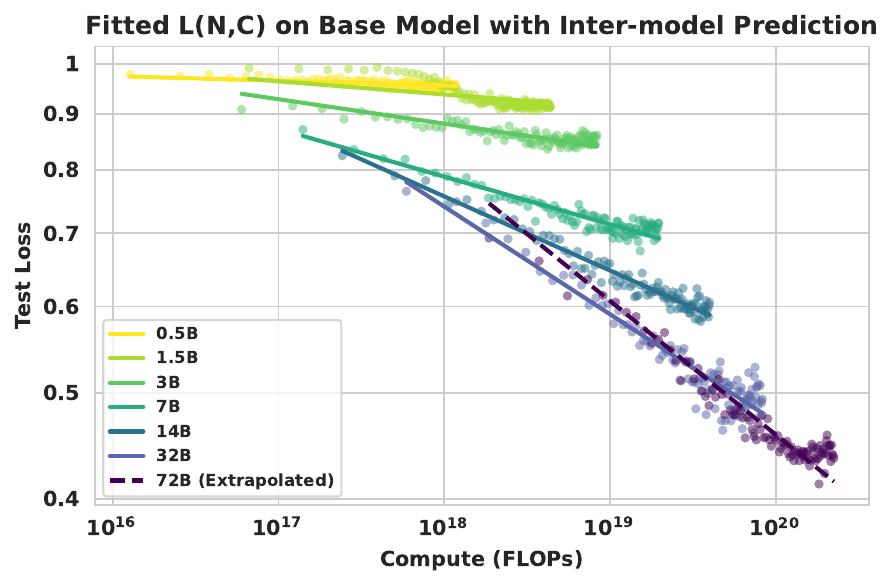}
        \caption{\tzl{Inter-model Prediction for Base Model}}
        \label{fig:loss_vs_compute_exp_a}
    \end{subfigure}
    % --- 子图1b ---
    \begin{subfigure}[b]{0.49\textwidth}
        \centering
        \includegraphics[width=\textwidth]{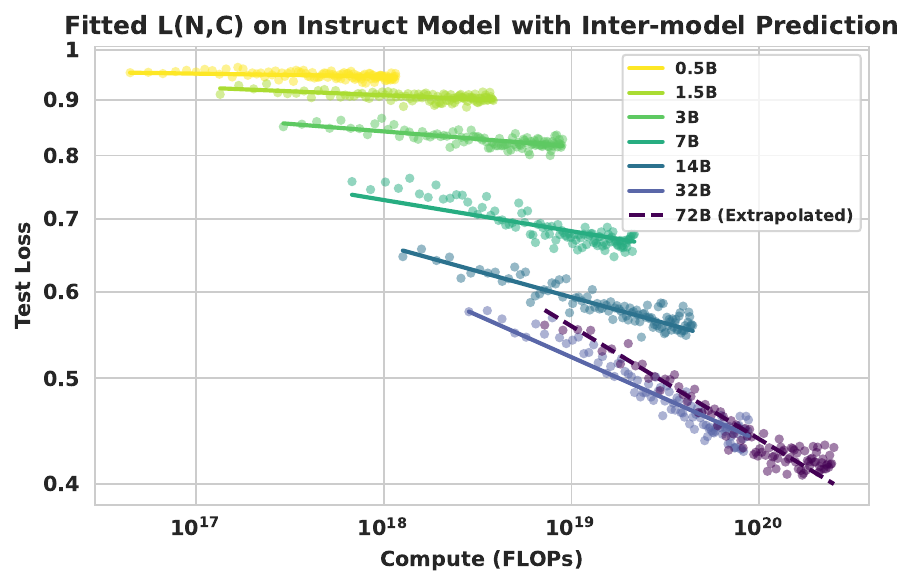}
        \caption{\tzl{Inter-model Prediction for Instruct Model}}
        \label{fig:loss_vs_compute_exp_b}
    \end{subfigure}
\end{figure*}

\begin{figure*}[t] % 保持位置参数一致
    \ContinuedFloat % <--- 关键命令：告诉LaTeX这是上一个图的延续，不要增加图号
    \centering
    
    % --- 子图1c ---
    \begin{subfigure}[b]{0.48\textwidth}
        \centering
        \includegraphics[width=\textwidth]{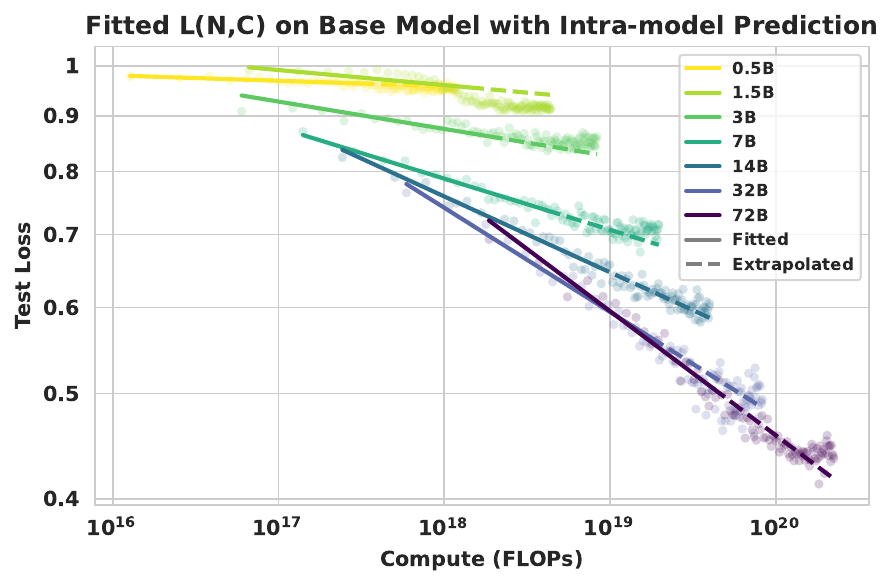}
        \caption{\tzl{Intra-model Prediction for Base Model}}
        \label{fig:loss_vs_compute_exp_c}
    \end{subfigure}
    % --- 子图1d ---
    \begin{subfigure}[b]{0.49\textwidth}
        \centering
        \includegraphics[width=\textwidth]{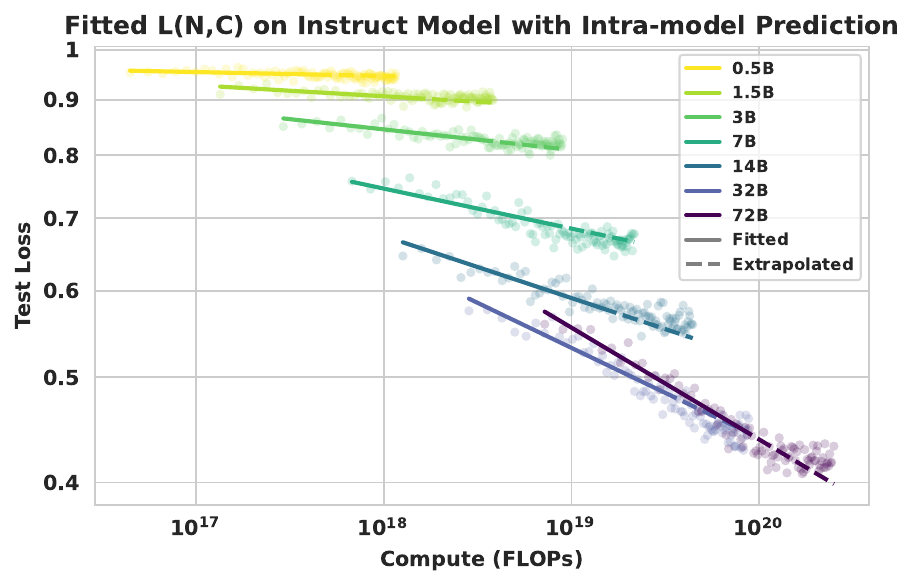}
        \caption{\tzl{Intra-model Prediction for Instruct Model}}
        \label{fig:loss_vs_compute_exp_d}
    \end{subfigure}
    
    % 第二部分的标题，通常写 "(cont.)" 或者不写
\caption{\textbf{Empirical Scaling Behaviors in RL Post-training.} 
We validate the proposed power-law (Eq.~\ref{eq:scaling_law_intro}), which characterizes the relationship between model performance and post-training resource consumption, across models ranging from 0.5B to 72B parameters.
\textbf{(a, b) Inter-model Prediction:} Scaling coefficients fitted on smaller models (0.5B--32B) accurately predict the learning efficiency of the held-out 72B model (dashed lines) .
\textbf{(c, d) Intra-model Prediction:} The final convergence on the full dataset is extrapolated solely from early training dynamics.}
    \label{fig:intro_scaling_prediction}
    \label{fig:loss_vs_compute_exp}
\end{figure*}
While these works have established foundational scaling principles for pretraining and RL in smaller neural networks, RL has become the dominant post-training strategy for enhancing LLMs' reasoning abilities—especially in mathematics, a domain requiring long-horizon, compositional reasoning~\citep{ferrag2025llm,guo2025deepseekr1, kimi2025scaling, ahn2024large}. However, despite its growing adoption for LLM reasoning tasks, a systematic understanding of how to effectively scale RL training remains elusive. In this work, we conduct a comprehensive empirical study to characterize these scaling behaviors across three critical resource regimes: (1) the \textbf{compute-constrained} scenario, where we determine the optimal model size to minimize test loss (\textbf{$1-PassRate$}) under a fixed FLOPs budget; (2) the \textbf{data-constrained} scenario, where we identify the model size yielding the lowest test loss with limited unique samples; and (3) the \textbf{data reuse} scenario, where we explore the trade-off between data uniqueness and reuse intensity under a fixed total data volume.

Based on our analysis across these regimes, we propose a \textbf{predictive formulation} that characterizes the relationship between test loss $L$, model size $N$, and resource budget $X$ (where $X$ denotes either Compute $C$ or Data $D$). We find that the scaling behavior can be effectively modeled by a power-law that exhibits a log-linear relationship between test loss and resource budget:
\begin{equation}
    \label{eq:scaling_law_intro}
    \log L(N, X) = -k(N) \cdot \log X + E(N)
\end{equation}
Here, $k(N)$ represents the learning efficiency, which our empirical results show does not grow indefinitely with increasing model scale. Instead, it follows a saturation trend modeled by:
\begin{equation}
    \label{eq:efficiency_intro}
    k(N) = \frac{K_{\text{max}}}{1 + \frac{N_{0}}{N}}
\end{equation}
This formulation (Eq.~\ref{eq:efficiency_intro}) shows that: while larger models consistently exhibit higher learning efficiency, the marginal gains in efficiency diminish as model size increases, asymptotically approaching a theoretical limit $K_{\text{max}}$.

To empirically validate this law and determine its parameters, \tzl{we fine-tune 63 LLMs with reinforcement learning on over 50k mathematics problems}, based on the Qwen2.5 model family~\citep{qwen2025qwen25technicalreport}. Figure~\ref{fig:loss_vs_compute_exp} shows that, \tzl{within the 0.5B-72B range}, the loss reduction brought by RL follows an approximately log-linear trend with compute. Importantly, larger models not only have better initial performance but also generally have more efficiency in computation and data utilization during the optimization process. Further analysis confirms that our proposed formulation (Eq.~\ref{eq:scaling_law_intro}) exhibits notable predictability, while also verifying the saturation effect in efficiency gains.

We additionally analyze the data-constrained regime, where we demonstrate that data reuse is a highly effective strategy. We validate the generality of our findings through extensive ablation studies on both base and instruct model series. Besides, we have also conducted experiments to study the impact of the rollout number in the GRPO algorithm~\citep{shao2024deepseekmathpushinglimitsmathematical}, shown in Appendix~\ref{subsec:ablation_grpo}. These investigations establish fundamental scaling relationships for RL post-training, providing a quantitative foundation and practical guidelines for resource-efficient model refinement.

Specifically, our key findings can be summarized as follows:
\vskip -0.2in
\begin{itemize}
    \item \textbf{We propose a predictive power-law formulation for RL post-training}, which models the scaling relationship between test loss, model size, compute, and data. This formulation enables reliable prediction of both larger model performance and late-stage training trajectories.
    \item Larger models consistently exhibit superior compute and data efficiency in the RL post-training, but \textbf{the marginal gains in efficiency diminish gradually with increasing model scale}, revealing an inherent saturation trend.
    \item In data-limited settings, repeated exposure to a small dataset is nearly as effective as using larger corpora, \textbf{highlighting data reuse as a practical strategy}.
\end{itemize}

\section{Experimental Setup}
\label{sec:experimental_setup}
We describe the experimental setup for studying scaling behavior in RL post-training of LLMs for mathematical reasoning, including the model family, training and evaluation data, and evaluation protocol in this section. Full details are provided in Appendix~\ref{app:exp_details}.

\paragraph{Models and Framework.} 
We use the Qwen2.5 model family~(0.5B, 1.5B, 3B, 7B, 14B, 32B and 72B parameters)~\citep{qwen2025qwen25technicalreport}, which shares the same architecture, so that parameter count is the only variable in our scaling analysis. All experiments are run with the VeRL framework~\citep{sheng2024hybridflow}, a large-scale RL platform for LLMs ensuring consistency and reproducibility.
\paragraph{Dataset settings.}The training data is the mathematics subset of the \texttt{guru-RL-92k} dataset from the Reasoning360 project~\citep{cheng2025revisiting}, which is carefully curated through deduplication and difficulty filtering. \tzl{We further sort the problems by increasing difficulty (decreasing pass rate, evaluated by Qwen2.5-7B-Instruct model) to enable curriculum learning}. The evaluation data consists of two parts. To verify proposed scaling law (Eq.~\ref{eq:scaling_law_intro}), we use a held-out set of 500 in-domain math problems sampled from the training distribution. To assess generalization, we evaluate on a broader benchmark suite spanning mathematics 
(AIME2024~\citep{patel2024aimeaioptimizationmultiple}, AMC2023~\citep{knoveleng_amc23_dataset}, GSM8K~\citep{cobbe2021trainingverifierssolvemath}, MATH500~\citep{lightman2023letsverifystepstep}), code (HumanEval~\citep{chen2021evaluatinglargelanguagemodels}), logic (Zebra Puzzle\citep{Lin_ZeroEval_A_Unified_2024}), and science (SuperGPQA~\citep{pteam2025supergpqascalingllmevaluation}). More details about dataset settings can be found in Appendix \ref{app:datasets}.

\paragraph{Prompt Setting.} To ensure stable behavior during RL training and evaluation, we use structured prompts tailored to each domain. For example, all mathematics problems are prepended with the Chain-of-Thought prompt~\citep{wei2023chainofthoughtpromptingelicitsreasoning}: \textit{``You are a knowledgeable math assistant. Answer the following questions and think step by step''}. More prompt templates for all related domains could be found in Appendix~\ref{app:prompts}.

\paragraph{RL Algorithm.}
We use Group Relative Policy Optimization~(GRPO)~\citep{shao2024deepseekmathpushinglimitsmathematical} for RL fine-tuning. GRPO estimates advantages by normalizing rewards across responses sampled from the same prompt, yielding a stable signal with lower memory cost. Specifically, for each question $q$, GRPO samples a group of outputs$\{o_1,o_2,\cdots,o_G\}$ from the old policy $\pi_{\theta_{old}}$, and the objective is defined as

\begin{equation}
\label{eq:GRPO}
\resizebox{\columnwidth}{!}{$
\begin{aligned}
\mathcal{L}_{\text{GRPO}}=
\frac{1}{G}\sum_{i=1}^{G}\frac{1}{\lvert o_i\rvert}\sum_{t=1}^{\lvert o_i\rvert}
\Bigl\{
  \min\Bigl[
  \rho(\theta)
      \,\hat A_{i,t},\; 
      \operatorname{clip}\!\Bigl(
          \rho(\theta),
         1-\varepsilon,\,
         1+\varepsilon
      \Bigr)\hat A_{i,t}
  \Bigr]
  -\beta\,\mathrm{D}_{\mathrm{KL}}
\Bigr\},
\end{aligned}
$}
\end{equation}

where $\rho(\theta)=\frac{\pi_{\theta}\!\left(o_{i,t}\mid q,\,o_{i,<t}\right)}{\pi_{\theta_{\mathrm{old}}}\!\left(o_{i,t}\mid q,\,o_{i,<t}\right)}$ is the important sampling weight. For each output $o_i$, a reward model or rule is used to yield the reward signal $\mathbf{r}=\{r_1,r_2,\cdots,r_G\}$. The advantage is computed as
\begin{equation}\label{eq:GRPO-Adv}
\hat{A}_{i,t} = \frac{r_i - \mathrm{mean}(\mathbf{r})}{\mathrm{std}(\mathbf{r})}.
\end{equation}

\paragraph{Reward Signal and Evaluation Metric.} For training, we use a binary reward signal in mathematical RL process, a reward of $1$ is given for a correct match and $0$ otherwise. For evaluation, we define our primary metric as the test loss ($L$), a proxy for reward-based performance in the RL setting. Formally, $L = 1 - (R / R_{\text{max}})$, where $R$ is the number of correct solutions and $R_{\text{max}}$ the total. \tzl{We adopt the term "test loss" for consistency with foundational neural scaling law literature (\cite{kaplan2020scaling})}. Notably, maximizing reward in RL training is equivalent to minimizing $L$.

\paragraph{Fitting and Prediction Protocols.}
To systematically evaluate the robustness and predictive capability of our derived scaling laws, we employ two distinct fitting protocols throughout our analysis:
\begin{itemize}
    \item \textbf{Inter-model Extrapolation:} We fit the scaling law parameters using data from smaller models (0.5B to 32B) to calculate the learning efficiency and predict the performance of the larger model (72B).
    \item \textbf{Intra-model Extrapolation:} We fit the scaling law using only the early training steps of a specific model to forecast its loss trajectory for the remainder of the training process. 
\end{itemize}

\section{Empirical Results and Scaling Laws}
\label{sec:experimental_analysis}

This section presents a comprehensive empirical investigation into the scaling behavior of RL for post-training LLMs. We first examine scaling behaviors under compute and data constraints, then analyze independent scaling dimensions, data reuse strategies, and finally evaluate generalization performance together. To ensure robust conclusions, each configuration is repeated \textbf{three times} for both base and instruct models ranging from 0.5B to 72B.

\subsection{Compute-Optimal Scaling}

\label{subsec:compute_scaling}
To characterize the scaling behavior under computational limits, we first formalize the Compute-Constrained Scenario. Given a fixed computational budget $C$, we seek to identify the optimal model size $N$ (and the corresponding data allocation $D$) that minimizes the final test loss. This can be expressed as the following constrained optimization problem:
\begin{equation}
\label{eq:compute-constrained}
    \underset{N,~D}{\argmin} \ L(N, D) \quad \text{s.t.}\quad \mathrm{FLOPs}(N,D) = C_{\text{const}},
\end{equation}
% \vskip 0.1in

To solve this, we train 0.5B–72B models and measure test loss as a function of cumulative FLOPs $C$. As shown in Figure~\ref{fig:loss_vs_compute_exp}, larger models consistently outperform smaller ones under the same compute budget for both base and instruct variants. \tzl{These plots include both Inter-model Extrapolation (fitted on 0.5B-32B and extrapolated on 72B) and Intra-model Prediction (predicting the remainder of training from initial steps) to demonstrate the predictive power of our derived scaling law.} The loss–compute relationship follows a log-linear trend, which can be modeled by a power-law:
\begin{equation}
\label{eq:compute_law}
\resizebox{\columnwidth}{!}{$
\begin{aligned}
    \log(L(N,C)) = -k_C(N) \cdot \log(C) + E_{C}(N), \\
    \ \text{where} \  k_C(N) = \left(\frac{K_{Cmax}}{1+\frac{N_{C}}{N}}\right)
\end{aligned}
$}
\end{equation}
To validate the predictive capability of Eq.~\ref{eq:compute_law}, we conducted two levels of evaluation under the protocols defined in Section \ref{sec:experimental_setup}. First, applying \textbf{Inter-model Extrapolation}, we fitted the scaling parameters on smaller models (0.5B–32B) to estimate the learning efficiency $k_C(N)$ for the 72B model. As illustrated in Figure \ref{fig:loss_vs_compute_exp_a} and \ref{fig:loss_vs_compute_exp_b}, the predicted efficiency aligns closely with the actual performance of the 72B model. Second, using \textbf{Intra-model Prediction}, we forecasted the remaining loss trajectory of specific models based solely on early training steps (Figure \ref{fig:loss_vs_compute_exp_c} and \ref{fig:loss_vs_compute_exp_d}), confirming the formula's robustness across different training stages.

\tzl{We further analyze learning efficiency term $k_C(N)$ in Eq.~\ref{eq:compute_law}. As Figure \ref{fig:coef_k_c} shows, $k_C(N)$ grows with model size $N$, meaning larger models consistently have higher learning efficiency. However, the efficiency gain from model scale is not uniformly linear. Beyond 32B, the increase in $k_C(N)$ diminishes, leading to efficiency saturation.} 

This saturation is manifested as a distinct performance crossover in Figure~\ref{fig:loss_vs_compute_exp}: In contrast to the immediate dominance of larger models in smaller parameter regimes, the 32B model outperforms the 72B counterpart initially under equivalent compute budgets, as the smaller model size inherently enables more training steps. We believe this observation reveals a latent trade-off between model scale and training steps in compute-constrained scenarios.

\subsection{Data-Optimal Scaling}
\label{subsec:data_scaling}
In many practical applications, the bottleneck lies not in compute but in the availability of high-quality reasoning data. We define the Data-Constrained Scenario as determining the model size $N$ that yields the lowest test loss given a limited amount of unique training data $D$:
\begin{equation}
    \label{eq:data-constraint}
    \underset{N,~C}{\argmin} \; L(N, C) \quad \text{s.t.} \quad D = D_{\text{const}}.
\end{equation}
% ==================================================================
% 第一幅图：Inter-model Prediction (跨模型预测)
% ==================================================================
% ==================================================================
% 第一组图：Inter-model Prediction (跨模型预测)
% ==================================================================
\begin{figure*}[t]
    \centering
    % --- 子图 (a) ---
    \begin{subfigure}[b]{0.48\textwidth}
        \centering
        \includegraphics[width=\linewidth]{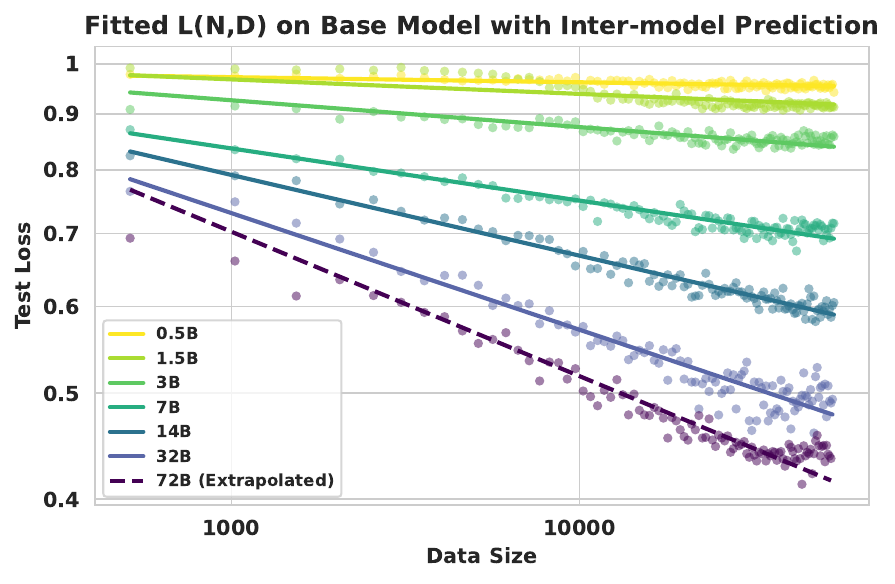}
        \caption{\tzl{Inter-model Prediction for Base Model}}
        \label{fig:inter_base}
    \end{subfigure}
    \hfill
    % --- 子图 (b) ---
    \begin{subfigure}[b]{0.48\textwidth}
        \centering
        \includegraphics[width=\linewidth]{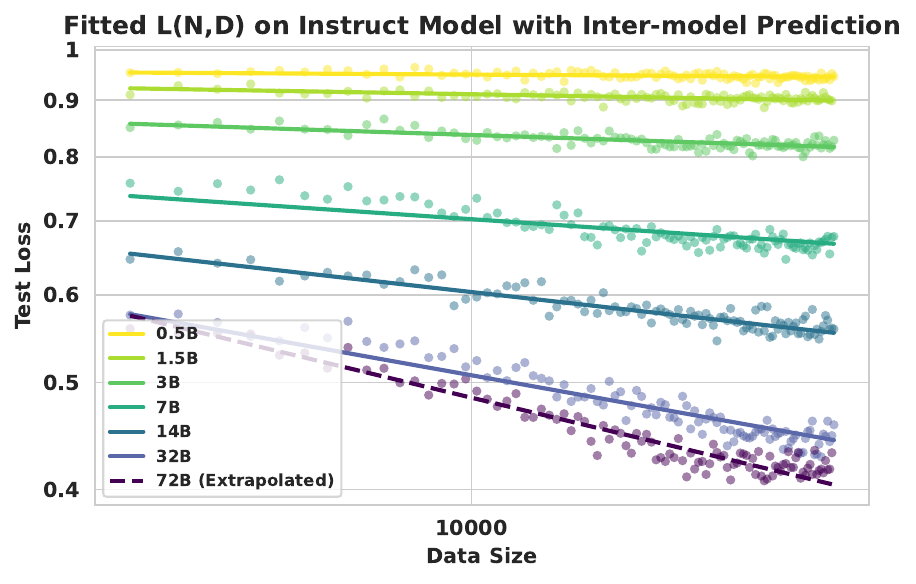}
        \caption{\tzl{Inter-model Prediction for Instruct Model}}
        \label{fig:inter_instruct}
    \end{subfigure}
    
    % --- Caption: 结合了预测方法 + 你指定的完整结论 ---
    \caption{\textbf{Inter-model Prediction in data scenario.} 
    The scaling law parameters are fitted on smaller models (0.5B--32B) to \textbf{predict} the learning efficiency of the largest model (72B, represented by dashed lines). 
    The accurate alignment validates that the superior sample efficiency of larger models follows a predictable trajectory, confirming that performance gains scale consistently up to 72B.}
    \label{fig:loss_vs_data_inter}
\end{figure*}

\begin{figure*}[b]
    \centering
    % --- 子图 (c) ---
    \begin{subfigure}[b]{0.48\textwidth}
        \centering
        \includegraphics[width=\linewidth]{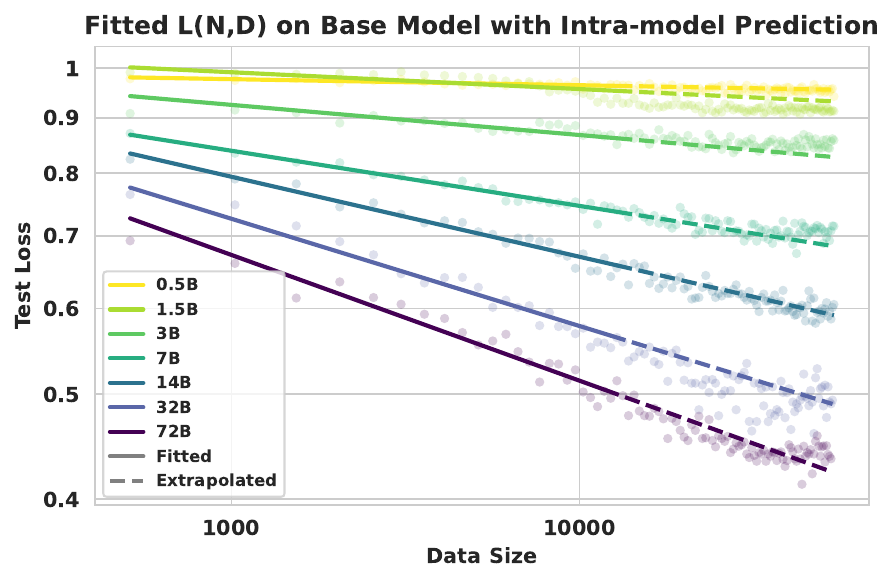}
        \caption{\tzl{Intra-model Prediction for Base Model}}
        \label{fig:intra_base}
    \end{subfigure}
    \hfill
    % --- 子图 (d) ---
    \begin{subfigure}[b]{0.48\textwidth}
        \centering
        \includegraphics[width=\linewidth]{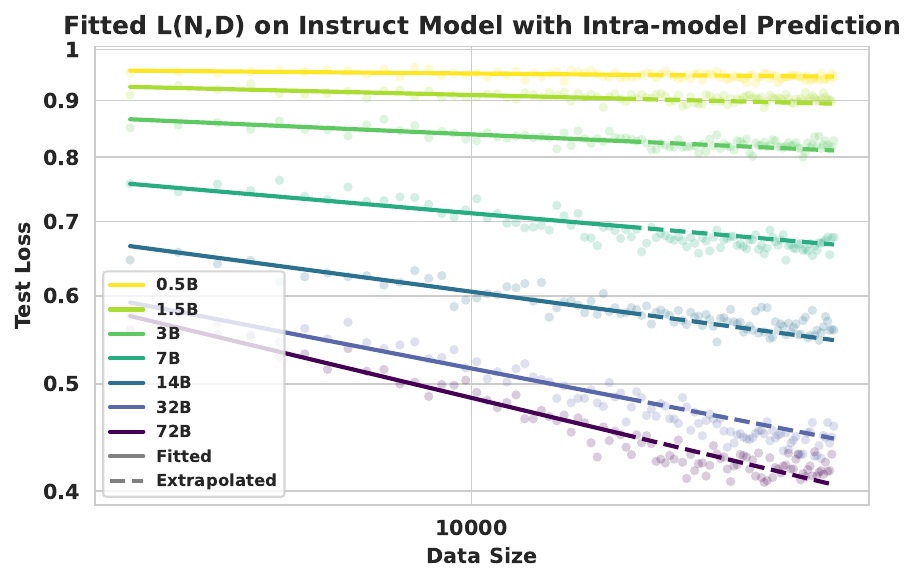}
        \caption{\tzl{Intra-model Prediction for Instruct Model}}
        \label{fig:intra_instruct}
    \end{subfigure}
    
    % --- Caption: 结合了预测方法 + 结论的可外推性 ---
    \caption{\textbf{Intra-model Prediction in data scenario.} 
    The loss trajectory is fitted using initial training data steps to \textbf{predict} the subsequent performance trends on the full dataset. 
    The results demonstrate that RL post-training maintains a nearly constant learning efficiency (manifesting as a linear trend in log-scale) during the rapid performance ascent, allowing the overall training trajectory to be accurately extrapolated from early dynamics.}
    \label{fig:loss_vs_data_intra}
\end{figure*}
\begin{figure*}[t] % 双栏排版通常使用 [t] 放在页顶，[H] 在 figure* 中通常无效
    \centering
    % --- 左图 ---
    \begin{subfigure}[b]{0.48\textwidth}
        \centering
        \includegraphics[width=\linewidth]{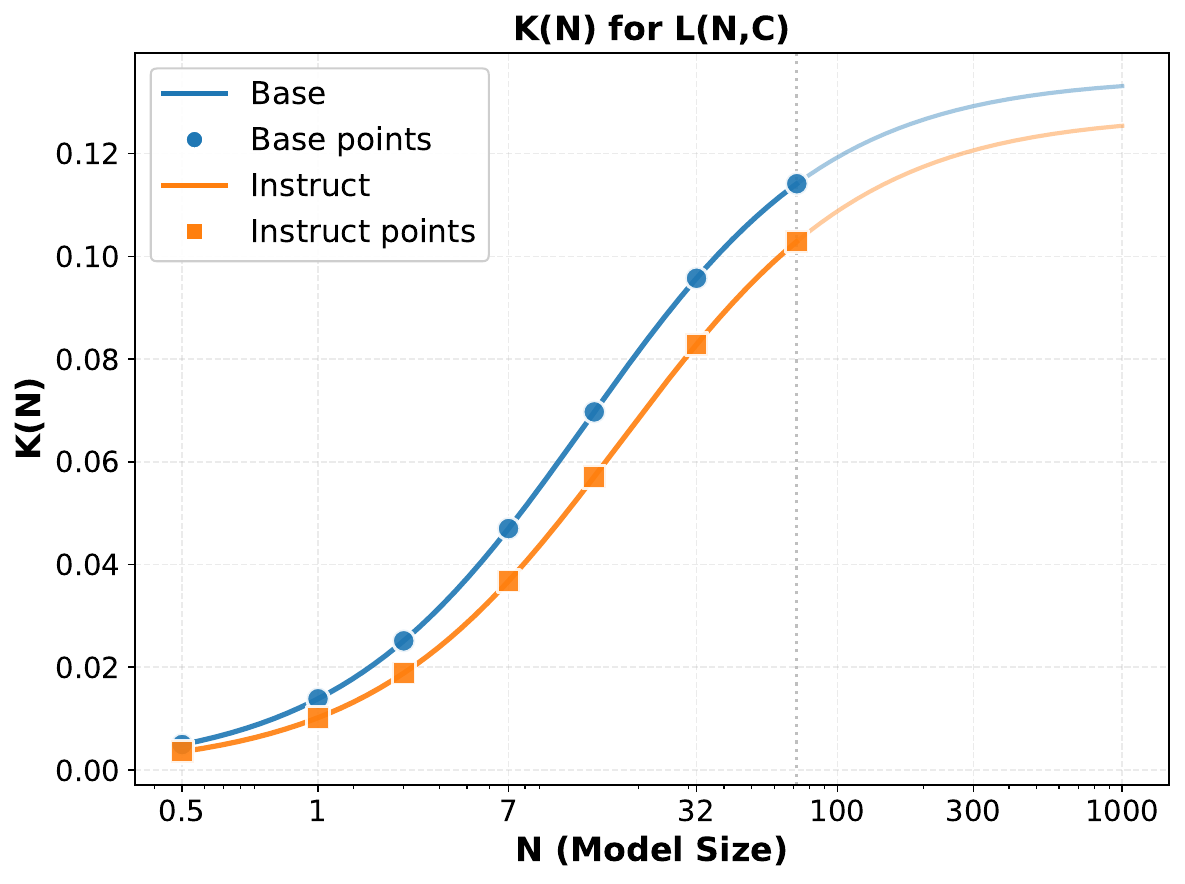}
        \caption{Fitted learning efficiency $k_C(N)$.} % 极简标题
        \label{fig:coef_k_c}
    \end{subfigure}
    \hfill % 左右留白
    % --- 右图 ---
    \begin{subfigure}[b]{0.49\textwidth}
        \centering
        \includegraphics[width=\linewidth]{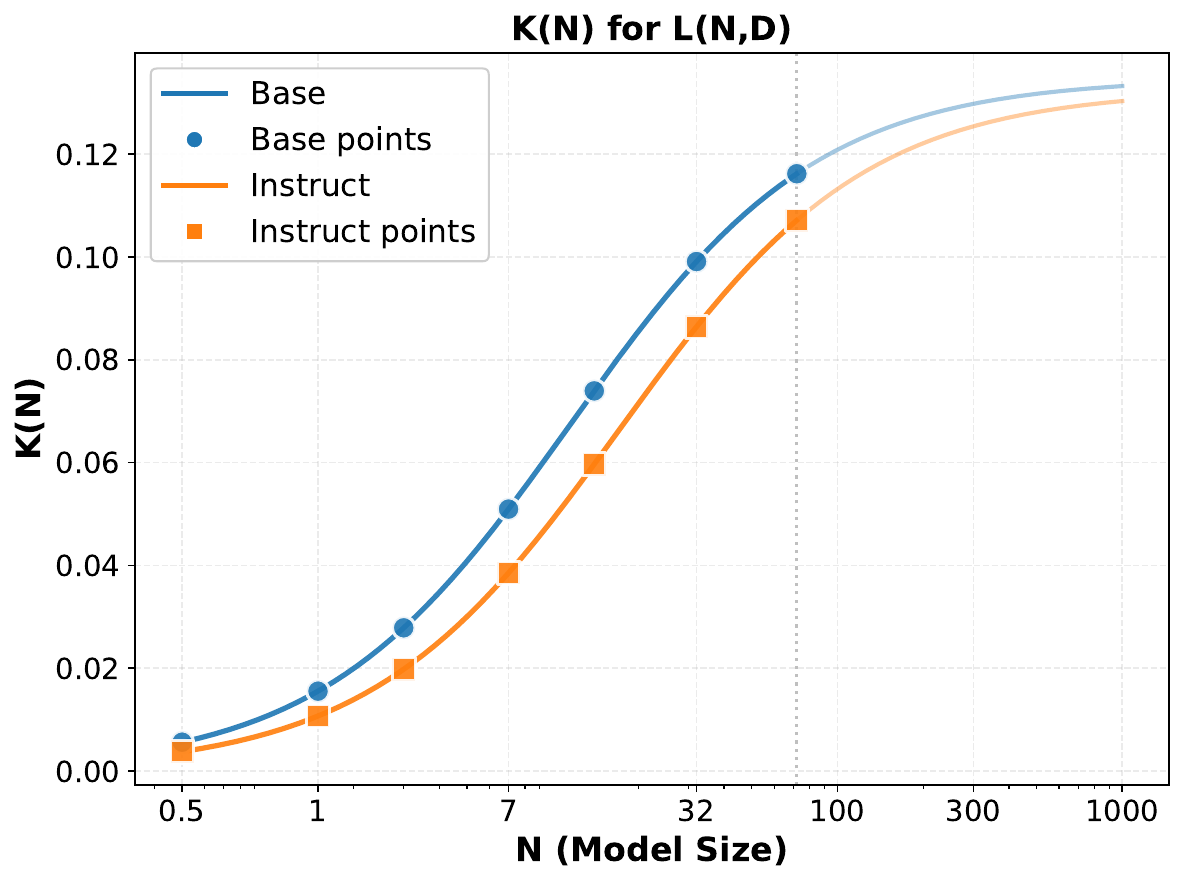}
        \caption{Fitted learning efficiency $k_D(N)$.} % 极简标题，长度一致
        \label{fig:coef_k_d}
    \end{subfigure}
    
    % --- 总标题 ---
    \caption{Fitted learning efficiency coefficients for Base and Instruct models. Both $k_C(N)$ (a) and $k_D(N)$ (b) exhibit identical trends: larger models consistently show higher learning efficiency, with efficiency gains beginning to diminish after the 32B model size.}
    \label{fig:coef_k_combined}
\end{figure*}

To empirically investigate this regime, we train models with varying parameter counts $N$ on fixed amounts of unique samples $D$. As shown in Figure~\ref{fig:loss_vs_data_inter} and Figure~\ref{fig:loss_vs_data_intra}, larger models consistently demonstrate superior sample efficiency within the 0.5B--72B range, achieving lower test loss under the same data constraints. This loss--data relationship also follows a log-linear trend and can be modeled by a formula analogous to the compute scenario:
\begin{equation}
\label{eq:data_law}
\resizebox{\columnwidth}{!}{$
\begin{aligned}
    \log(L(N,D)) = -k_D(N) \cdot \log(D) + E_{D}(N),\\
    \text{where} \ \  k_D(N) = \left(\frac{K_{Dmax}}{1+\frac{N_{D}}{N}}\right)
\end{aligned}
$}
\end{equation}
Mirroring the analysis in Section~\ref{subsec:compute_scaling}, we evaluate the predictive capability of our data scaling law (Eq.~\ref{eq:data_law}) in two consistent settings: inter-model extrapolation (Shown in Figures \ref{fig:loss_vs_data_inter}) and intra-model prediction (Shown in Figures \ref{fig:loss_vs_data_intra}), and the predictions also align with actual results closely.
Moreover, the data learning efficiency $k_{D}(N)$ (Figure~\ref{fig:coef_k_d}) exhibits the same saturation trend as $k_{C}(N)$ (Figure~\ref{fig:coef_k_c}), where marginal efficiency gains diminish significantly beyond 32B.

The unified functional form across both compute and data domains underscores the theoretical consistency of our scaling law. Quantitative analysis further validates this model, achieving a high goodness-of-fit ($R^2 > 0.99$) across both intra- and inter-model predictions. The detailed fitted parameters, including the saturation scale $N_0$ and efficiency limit $K_{max}$, are reported in Table~\ref{tab:kn_parameters} in Appendix~\ref{app:fitting_4_KE}.

\subsection{Cross-Architecture Generalization}
\label{subsec:cross_arch}

\begin{figure*}[b]
    \centering
    \begin{subfigure}[b]{0.48\textwidth}
        \centering
        \includegraphics[width=\linewidth]{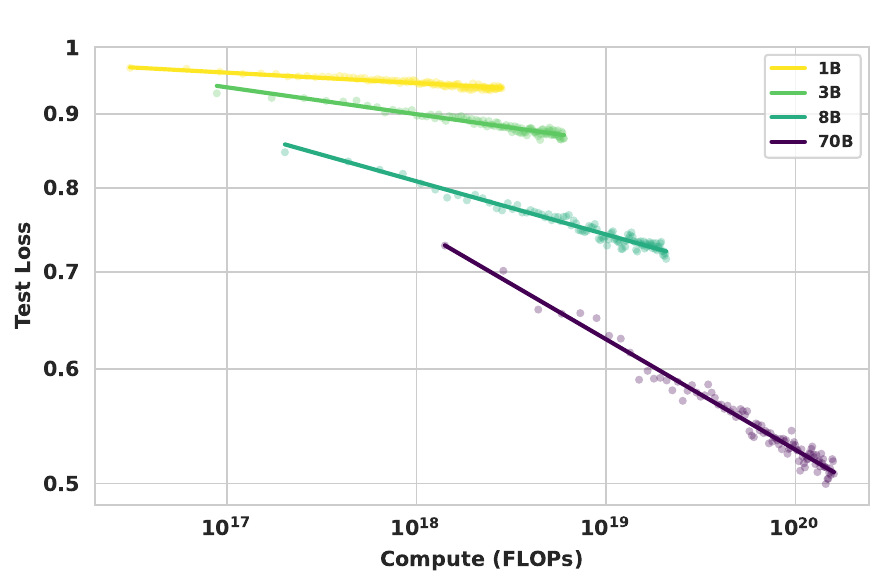}
        \caption{Fitted $L(N, C)$ on Llama Instruct}
        \label{fig:llama_scaling_c}
    \end{subfigure}
    \hfill
    \begin{subfigure}[b]{0.48\textwidth}
        \centering
        \includegraphics[width=\linewidth]{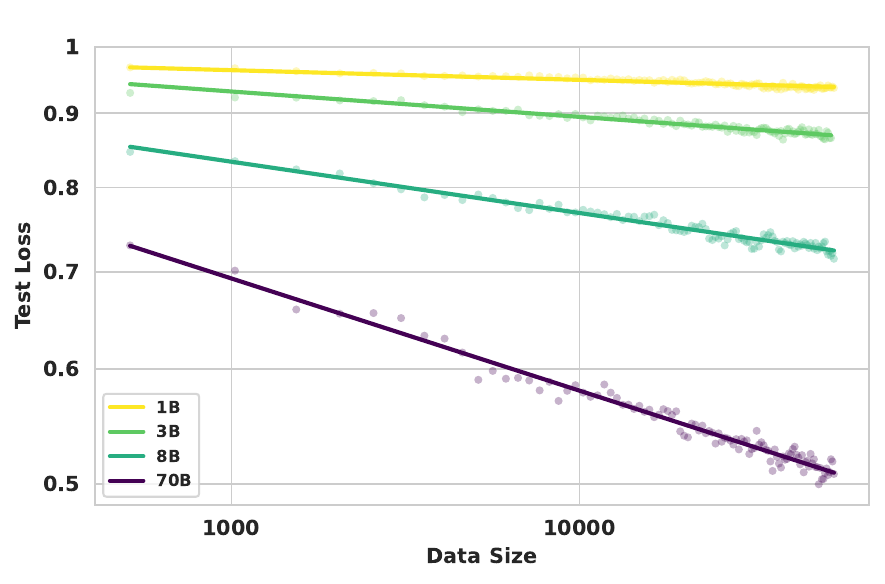}
        \caption{Fitted $L(N, D)$ on Llama Instruct}
        \label{fig:llama_scaling_d}
    \end{subfigure}
    \caption{\textbf{Cross-architecture validation on Llama~3.} The same scaling law (Eq.~\ref{eq:compute_law},~\ref{eq:data_law}) fitted on Llama~3 Instruct models (1B--70B) achieves $R^2 > 0.99$ for both compute (a) and data (b) dimensions.}
    \label{fig:llama_scaling}
\end{figure*}

All experiments in Sections~\ref{subsec:compute_scaling} and~\ref{subsec:data_scaling} are conducted on the Qwen2.5 model family. A natural question is whether the observed scaling behavior is specific to a particular architecture or represents a more general property of RL post-training. To investigate this, we apply the same experimental protocol to the Llama~3 model family, spanning 1B to 70B parameters (Llama-3.2-1B/3B-Instruct and Llama-3.1-8B/70B-Instruct).

We train each model using identical GRPO configurations and evaluate on the same holdout benchmark. As shown in Figure~\ref{fig:llama_scaling}, both the compute scaling law $L(N,C)$ (Eq.~\ref{eq:compute_law}) and the data scaling law $L(N,D)$ (Eq.~\ref{eq:data_law}) fit the Llama results with $R^2 > 0.99$, confirming that the proposed log-linear formulation and the saturation structure of $k(N)$ generalize across architectures. Notably, while Llama models achieve lower absolute performance than Qwen at comparable sizes (e.g., Llama-70B peaks at ${\sim}50\%$ holdout accuracy vs.\ Qwen-72B at ${\sim}59\%$), the \emph{functional form} of the scaling relationship remains identical. This suggests that the scaling dynamics of RL post-training are governed by the optimization process itself, rather than by architecture-specific properties of the base model.

\subsection{Scaling with Constrained Data and Reuse}
\begin{figure}[h]
    \centering
    \includegraphics[width=0.75\linewidth]{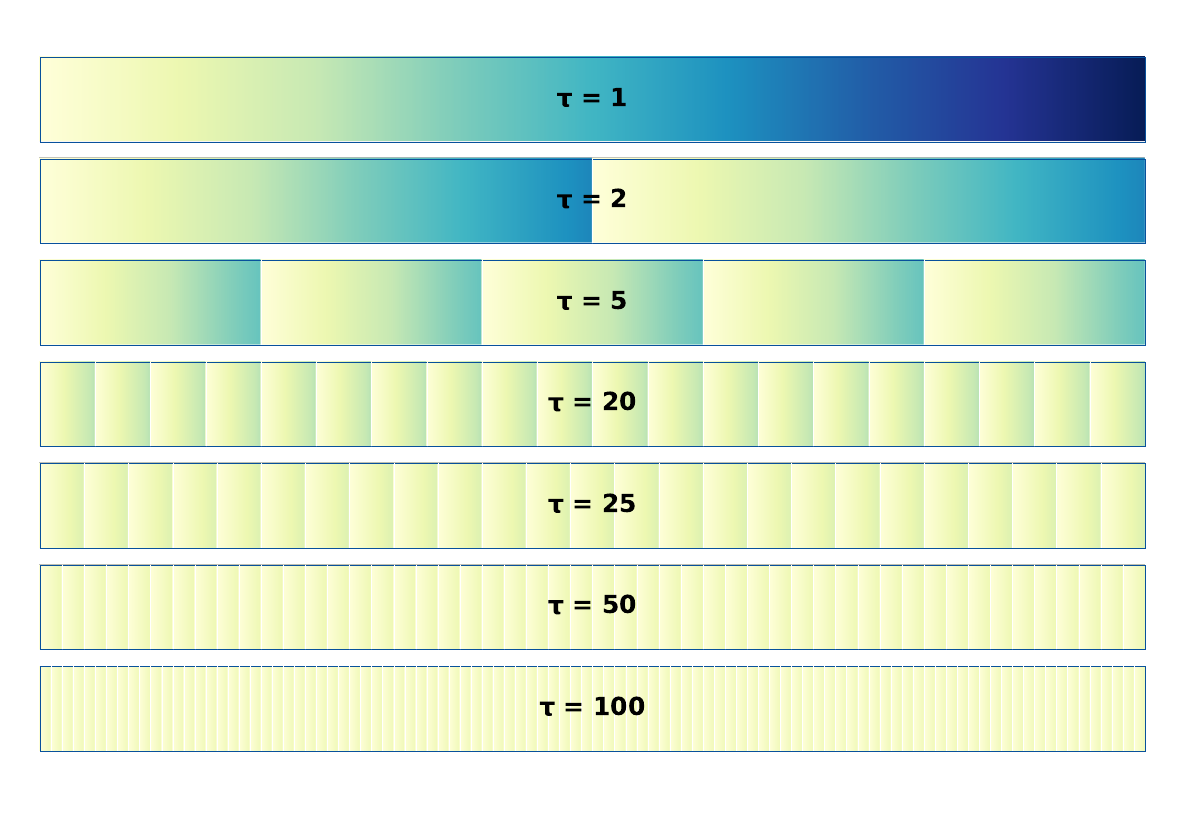}
    \caption{Data reuse schema. Each row represents a training run with reuse factor $\tau$. All runs use the same total data volume, but with varying subset sizes and repetition counts.}
    \label{fig:data_reuse_schema}
\end{figure}

When unique data is scarce, a critical question is whether repeating data is effective. We investigate this Data Reuse Scenario by fixing the total data budget and varying the reuse factor $\tau$. Specifically, we aim to identify the optimal reuse factor $\tau$ that minimizes the test loss:
\begin{equation}
\underset{\tau}{\text{argmin}} ; L(\tau) \quad \text{s.t.} \quad D_{\text{unique}} \times \tau = D_{\text{total}},\label{eq:data-reuse}
\end{equation}
\label{subsec:data_reuse}
To systematically evaluate this problem, we simulate data constraints by partitioning the training set into smaller subsets while preserving the difficulty distribution (Details provided in Appendix~\ref{app:data-reuse-detailed}). Each subset is cycled through multiple times, with $\tau$ controlling the repetition frequency. As illustrated in Figure~\ref{fig:data_reuse_schema}, different reuse factors correspond to different subset sizes, all producing the same total data volume. Crucially, $D_{\text{total}}$ is kept fixed across runs, and curriculum ordering is maintained to ensure that performance differences arise solely from the degree of data reuse rather than distributional artifacts.
\begin{figure}[t]
    \centering
    \includegraphics[width=\linewidth]{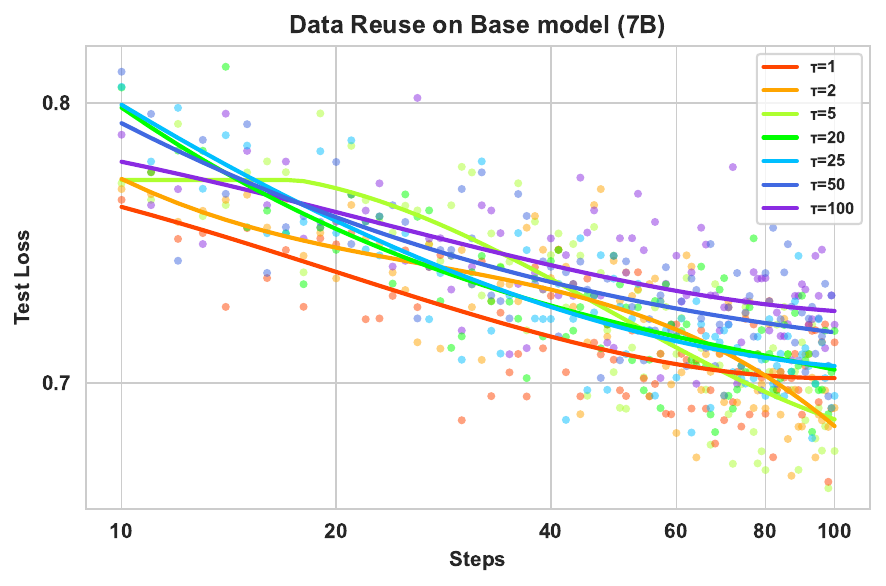}
    \caption{Performance in data-constrained settings is primarily determined by $D_{\text{total}}$. For a fixed $D_{\text{total}}$, the final test loss is insensitive to $\tau$, with no significant degradation up to $\tau=25$.}
    \label{fig:data_reuse_ablation}
\end{figure}

Under this experimental setup, our results demonstrate that performance is primarily governed by the total number of optimization steps ($D_{\text{total}}$), rather than sample uniqueness. As illustrated in Figure~\ref{fig:data_reuse_ablation} (for instruct model, check Figure~\ref{fig:data_reuse_instruct} in Appendix~\ref{app:data-reuse-detailed}), the final test loss proves insensitive to the reuse factor, showing no significant degradation for $\tau \leq 25$. However, the limit exists at $\tau=100$, we observe clear signs of overfitting, indicating that excessive repetition eventually harms generalization. Collectively, these findings confirm that moderate data reuse is a highly effective strategy for RL fine-tuning in data-constrained regimes.

\subsection{Domain Transfer}
\label{subsec:domain-transfer}
\begin{figure*}[t]
    \centering
    % --- Left Subfigure (In-domain) ---
    \includegraphics[width=\textwidth]{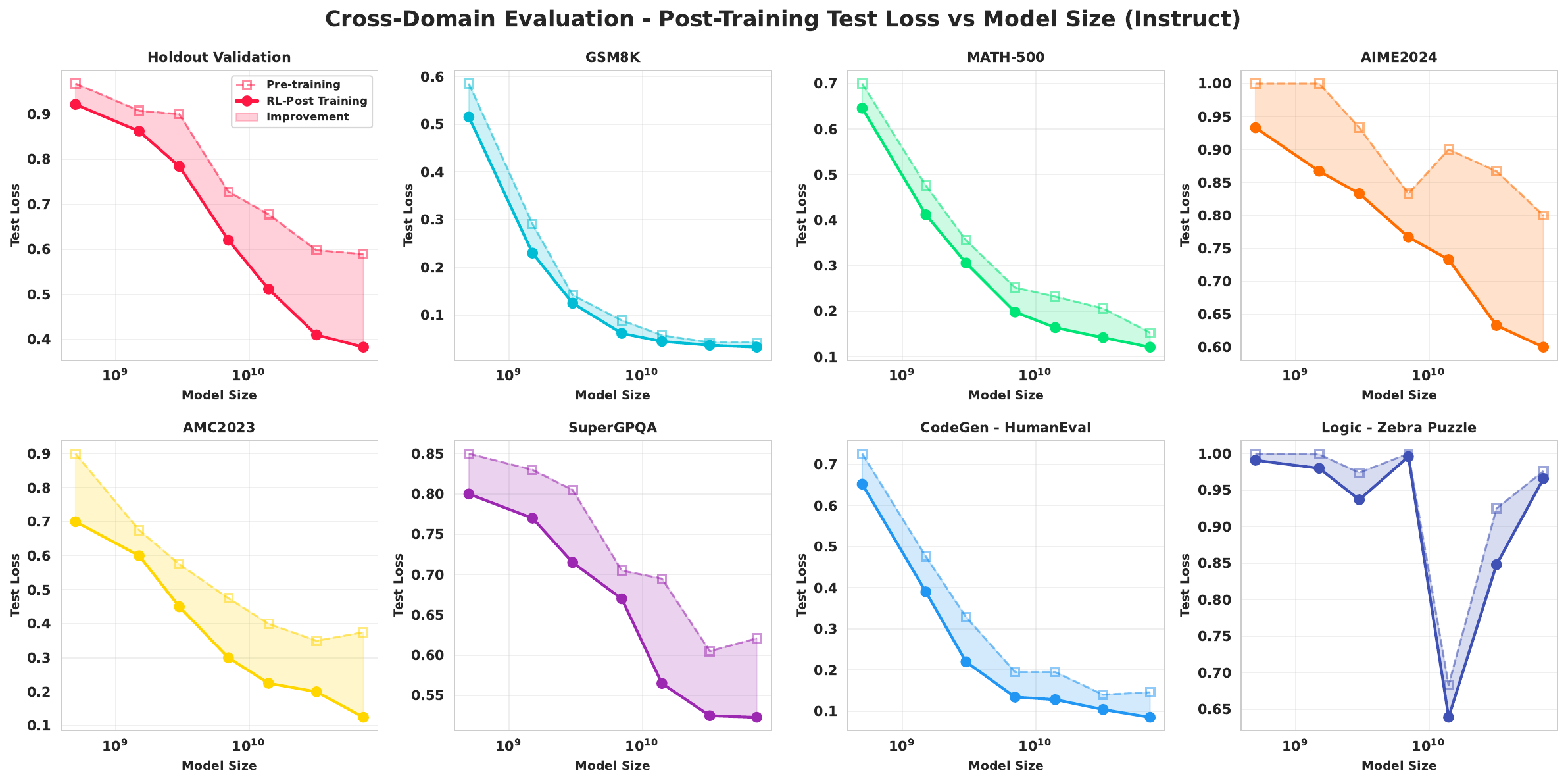}
    \caption{\tzl{RL post-training on mathematical reasoning yields generalization improvements on in-domain tasks with varying difficulty, but shows negligible transfer to out-of-domain tasks.}}
    \label{fig:domain_transfer}
\end{figure*}
We investigate the generalization capabilities of reinforcement learning fine-tuning (RFT) by evaluating models on a comprehensive suite of in-domain and out-of-domain (OOD) benchmarks. Our results demonstrate a divergent trends: while RL post-training yields robust generalization improvements on in-domain mathematical tasks of varying difficulty, it shows negligible transfer to out-of-domain tasks. (Detailed results are provided in Appendix~\ref{app:generalization_tasks}).

\textbf{In-Domain Generalization.} Figure~\ref{fig:domain_transfer} shows consistent improvements on unseen mathematics tasks outside the training set. On benchmarks, from easy to hard, including \texttt{GSM8K}, \texttt{MATH-500}, \texttt{AMC2023}, \texttt{AIME2024}, test loss steadily decreases with training compute, suggesting that RL post-training enhances transferable reasoning skills within mathematics.

\textbf{Out-of-Domain Generalization.} As shown in Figure~\ref{fig:domain_transfer}, results on OOD tasks are markedly different. For code generation (\texttt{HumanEval}) and STEM problems (\texttt{SuperGPQA}), performance gains marginally, indicating that RL fine-tuning is highly specialized.\tzl{On logical reasoning (\texttt{zebra\_puzzle}), performance degrades for larger models, suggesting that intensive optimization on mathematical reasoning may interfere with or "damage" other distinct reasoning abilities}.

\section{Related Work}
\textbf{Foundational Scaling Laws of Neural Language Models.} Foundational scaling studies show language modeling loss follows smooth power-laws in model size $N$, data $D$, and compute $C$ \citep{kaplan2020scaling}, with compute-optimal training prescribing near lockstep growth of parameters and tokens under fixed FLOPs \citep{hoffmann2022training}. Later analyses attribute earlier discrepancies to embedding/non-embedding parameter accounting, last-layer costs, optimizer warmup, and scale-sensitive hyperparameters \citep{pearce2024reconciling,porian2024resolving}, while data-centric refinements examine pruning efficiency \citep{sorscher2022beyond}, repetition effects \citep{hernandez2022scaling}, gzip-based complexity predictors \citep{pandey2024gzip}, constrained or synthetic regimes \citep{muennighoff2023scaling,qin2025scaling}, and task transfer (e.g., translation) \citep{isik2024scaling}. Test-time compute amplification supplies an inference analogue to classical training laws \citep{snell2024scaling}. 

\textbf{RL post-training in LLMs.} In RL, power-law trends similarly link capacity, interaction compute, and performance \citep{hilton2023scaling}; scaling RFT across horizon and compute improves mathematical and coding reasoning \citep{guo2025deepseekr1,kimi2025scaling,mai2025agent,zhang2025landscapeagenticreinforcementlearning,zhang2025surveyreinforcementlearninglarge}, while extended schedules \citep{liu2025prorl}, ultra-low-shot or single-example RL \citep{wang2025reinforcement}, and minimal-data efficiency paradigms \citep{li2025limr} probe data–compute tradeoffs. Instability and uneven gains highlight fragile optimization \citep{zeng2025simplerlzoo,yue2025doesrlincentivize}, and multi-domain mixtures reveal both synergy and interference across math, code, and logic \citep{li2025onedomainhelps,cheng2025revisiting}. Recent work further advances RL-based reasoning through attention-guided process supervision~\citep{nie2026attnpo}, self-search reinforcement learning~\citep{fan2025ssrl}, and structured analysis of LLM reasoning behaviors~\citep{yuan2025understandingllmreasoningabstractive}.

\textbf{Mathematical Reasoning with LLMs.} Mathematical reasoning amplifies these dynamics: accuracy generally scales upward while verification behaviors remain inconsistent \citep{touvron2023scaling}; corpus volume and quality jointly shape attainable curves \citep{ye2024skywork}; multi-task math-generalist training diverges from specialist scaling trajectories \citep{yue2023mammoth}; and RL with code execution induces additional behaviors such as emergent tool use concentrated in math problem solving \citep{zeng2025agent}. Collectively, evidence indicates that reasoning performance is governed by interacting axes of model size, data distribution/quality, training (supervised vs.\ RL) paradigm, and allocation of both training and inference compute, while unified laws for mathematical reasoning remain only partially characterized.
\section{Discussion}
\textbf{Scaling Dependence on Evaluation Environment and Metrics.} 
Reinforcement learning optimizes directly for environment rewards~\citep{10.5555/3312046}, which in principle allows unbounded capability—as demonstrated by AlphaZero mastering board games~\citep{silver2017masteringchessshogiselfplay}, AlphaFold predicting protein structures~\citep{jumper2021highly}, and frontier LLMs such as Gemini-2.5-Pro achieving IMO-level performance~\citep{huang2025gemini25procapable}. In contrast, text-based LLMs lack a well-defined RL environment, forcing us to rely on human-curated datasets as proxies. Test loss thus serves as a pragmatic but imperfect metric: it is monotonic and convergent, yet heavily dependent on dataset construction and task difficulty, with different benchmarks (e.g., GSM8K vs. AIME, Section~\ref{subsec:domain-transfer}) showing distinct convergence rates. This task dependence makes the absolute coefficients of our fitted scaling laws ($K_{max}, N_0, E$) difficult to interpret universally. Prior work proposed ``intrinsic performance’’—the minimum compute needed to reach a target reward—as a normalization across environments~\citep{hilton2023scaling}, but we did not find an analogous measure in large-scale LLMs. Establishing principled, environment-independent evaluation protocols remains an open and critical challenge for RL-based scaling studies.

\textbf{Scaling Dependence on Model Scale.} 
While larger models (0.5B-72B) exhibit superior efficiency, our analytic term $k(N)$ confirms this advantage is not infinite. Instead, efficiency gains follow a saturation curve toward a theoretical limit $K_{max}$ (Eq.~\ref{eq:efficiency_intro}), implying diminishing marginal returns at extreme scales.
This finding implies that scaling up models beyond a certain point, while still yielding absolute performance gains, suffers from diminishing marginal returns in efficiency.
% Our study examined models from 0.5B to 14B parameters, revealing consistent gains in sample and compute efficiency for larger models under RL post-training. This range, however, is still below frontier deployments, and our larger-scale experiments are underway. We anticipate that the observed advantages hold when tasks are of moderate difficulty and data are well curated, but may not extend indefinitely. On simple benchmarks such as GSM8K, performance tends to saturate as models grow, while for extremely hard problems (e.g., NP-hard tasks or open research challenges) neither small nor large models exhibit meaningful convergence. A plausible hypothesis is the existence of a critical model scale: large enough to encode broad commonsense knowledge for efficient RL adaptation, but not so over-parameterized that returns diminish. Identifying such a “sweet spot’’ remains an important direction for future work.

\textbf{Dependence on RL Algorithm.}
Our analysis is based on GRPO, a mainstream and stable RL post-training algorithm for LLMs that uses an actor-only design and normalizes rewards across responses. Comparative study with alternative RL algorithms ~\citep{cui2025entropymechanismreinforcementlearning} reports minor differences in training curves. Whether more advanced algorithms can significantly improve sample efficiency or stability—and thereby reshape the scaling frontier—remains an important open question.

\textbf{Future of LLM Agent.}
The integration of reinforcement learning with agentic LLMs is increasingly viewed as a promising direction~\citep{zhang2025landscapeagenticreinforcementlearning, zhang2025surveyreinforcementlearninglarge}. Both theoretical and empirical studies show that augmentations such as external tool use and long-term memory can substantially boost model performance~\citep{lin2025understandingtoolintegratedreasoning,houliston2025provablebenefitsintoollearning,mai2025agentrlscalinglaw}. In the multi-agent setting, reward-driven self-organization~\citep{zhou2025reso} and co-evolutionary interaction rewards~\citep{xue2026comascoevolvingmultiagentsystems} further show that collaborative LLM agents can surpass individual model performance. We anticipate that such agentic mechanisms will markedly improve the scaling behavior of RL-trained LLMs: by offloading deterministic computations to tools and focusing learning on high-level decision making, these models could achieve much higher efficiency, effectively shifting the performance frontier upward for a given compute or data budget. Understanding the scaling laws of these agentic systems is, therefore, a key and exciting avenue for future research.

\section{Conclusion}
In this work, we conducted a comprehensive empirical study to characterize the scaling behaviors of reinforcement learning post-training for large language models. Our analysis yields three fundamental findings: 

First, we establish a predictive power-law formulation that models the log-linear relationship between test loss and resource consumption. We demonstrate that this formulation enables reliable performance forecasting, accurately predicting the learning efficiency of larger model and estimating final trajectories from early training dynamics. 

Second, we identify a inherent saturation trend in learning efficiency. While larger models consistently exhibit superior compute and sample efficiency, the marginal gains in the learning efficiency coefficient $k(N)$ do not grow indefinitely; instead, they diminish as model size increases, asymptotically approaching a theoretical limit. 

Third, we find that RL model performance in data-constrained regimes is primarily governed by the total volume of training data rather than sample uniqueness. Our experiments validate that moderate data reuse is a highly effective strategy, where increasing the repetition factor allows models to maintain performance improvements without requiring additional unique samples.

We further validate the generality of these findings on the Llama~3 family, confirming the scaling law is architecture-agnostic.
\newpage
\section*{Limitations}
First, our conclusions are currently limited to reinforcement learning within the mathematical domain and have not yet been extended to multi-domain RL scenarios. Second, constrained by the Qwen2.5 series limit of 72B parameters, we could not empirically verify the efficiency saturation trend on larger-scale models beyond this size. Finally, our experiments focus on dense model families (Qwen2.5 and Llama~3); the generalizability to Mixture-of-Experts (MoE) architectures remains to be explored.

\section*{Acknowledgments}
This research was supported by the National Natural Science Foundation of China (No.\ 62506186). This work is supported by Shanghai Artificial Intelligence Laboratory.

\bibliography{custom}

\appendix
\label{sec:appendix}

\section{Experiment Setup Details}
\label{app:exp_details}

This section provides a detailed breakdown of the datasets and hyperparameters used in our experiments.
\subsection{Dataset Details}
\label{app:datasets}
Our training was conducted on a curated mathematics dataset. For evaluation, especially for analyzing generalization (as mentioned in the main text), we utilized a comprehensive suite of benchmarks spanning multiple domains. The composition of this evaluation suite is detailed in Table~\ref{tab:eval_suite_appendix}.

\begin{table*}[t]
\centering
\begin{tabular}{lrll}
\toprule
\textbf{Dataset} & \textbf{Samples} & \textbf{Huggingface Tag} & \textbf{Domain} \\
\midrule
Held-out Data & 500 & LLM360/guru-RL-92k & Math \\
aime2024 & 30 & Maxwell-Jia/AIME\_2024 & Math \\
amc2023 & 40 & knoveleng/AMC-23 & Math \\
codegen\_humaneval & 164 & openai/openai\_humaneval & Code \\
gsm8k & 1319 & openai/gsm8k & Math \\
logic\_zebra\_puzzle & 200 & LLM360/guru-RL-92k & Logical Reasoning \\
math & 500 & HuggingFaceH4/MATH-500 & Math \\
stem\_supergpqa & 200 & LLM360/guru-RL-92k & STEM \\
\midrule
\textbf{Total} & \textbf{2953} & & \\
\bottomrule
\end{tabular}
\caption{Composition of the multi-domain evaluation suite.}
\label{tab:eval_suite_appendix}
\end{table*}

\subsection{Hyperparameter Configuration}
\label{app:hyperparams}
All experiments were conducted with a consistent set of hyperparameters for the Group Relative Policy Optimization (GRPO) algorithm to ensure a fair comparison across different model sizes and configurations. The key hyperparameters are listed in Table~\ref{tab:hyperparams_appendix}.

\begin{table}[H]
\centering
\begin{tabular}{ll}
\toprule
\textbf{Hyperparameter} & \textbf{Value} \\
\midrule
Learning Rate & $1e-6$ \\
Batch Size & 512 \\
KL Loss Coefficient & 0.001 \\
Rollout Temperature (Training) & 1.0 \\
Rollout Temperature (Evaluation) & 0.7 \\
Clip Ratio (High \& Low) & 0.2 \\
Input Sequence Length & 2048 \\
Output Sequence Length & 4096 \\
\bottomrule
\end{tabular}
\caption{GRPO training hyperparameters used across all experiments.}
\label{tab:hyperparams_appendix}
\end{table}

\subsection{Prompt Templates}
\label{app:prompts}
This section details the specific prompt templates used for evaluating models on different domains. For each task, the model was provided with the corresponding instruction prepended to the problem statement \texttt{<question>}. Check the details in Table~\ref{tab:prompt_templates}.
\begin{table*}[tb]
\centering
\begin{tabular}{>{\raggedright\arraybackslash}p{3cm} >{\raggedright\arraybackslash}p{\dimexpr\linewidth-3cm-2\tabcolsep\relax}}
\toprule
\textbf{Domain} & \textbf{Prompt Template} \\
\midrule
Mathematics & \texttt{You are a knowledgeable math assistant. Answer the following questions and think step by step\textbackslash{}n<question>\textbackslash{}nPlease output the final answer within \textbackslash{}\textbackslash{}boxed\{\}.} \\
\midrule
Code & \texttt{Write a complete, self-contained Python solution to the following problem. Your solution must include all necessary imports and the full function definition, including the signature exactly as specified. Do not modify the function signature or docstring.\textbackslash{}n<question>} \\
\midrule
Logic & \texttt{Solve the following puzzle\textbackslash{}n<question>\textbackslash{}nPlease return the final answer in <answer> </answer> tags, for example <answer> \{"header": ["Position", "Nationality", "Job"], "rows": [["1", "british", "plumber"], ["2", "polish", "carpenter"]]\} </answer>.}  \\
\midrule
Science (STEM) & \texttt{You are a knowledgeable assistant. Answer the following questions and think step by step \textbackslash{}n <question> \textbackslash{}n put your final answer option within \textbackslash{}\textbackslash{}boxed\{\}. Only put the letter in the box, e.g. \textbackslash{}\textbackslash{}boxed\{A\}. There is only one correct answer} \\
\midrule
\end{tabular}
\caption{Prompt templates used for different evaluation domains.}
\label{tab:prompt_templates}
\end{table*}

\subsection{Data Reuse Experiment Setup}
\label{app:data-reuse-detailed}

\begin{figure}[H]
    \centering
    % Placeholder: Instruct model's Final Loss vs. Data Reuse Factor
    \includegraphics[width=\linewidth]{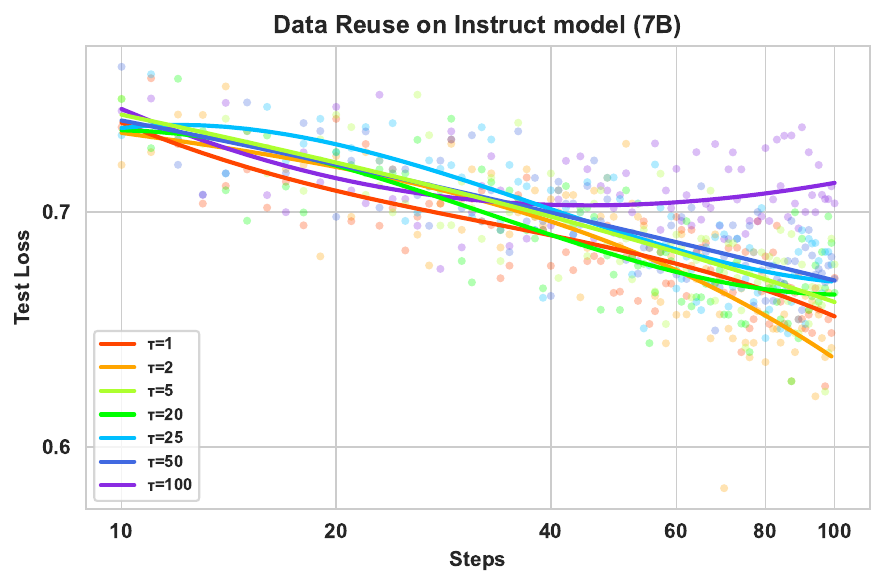}
    \caption{Instruct Model}
    \label{fig:data_reuse_instruct}
\end{figure}

To systematically evaluate the effect of data reuse under constrained data scenarios, we design controlled experiments where all runs are trained with the same total data size but different levels of data repetition. Each run randomly samples a subset from the full training corpus and repeats this subset sufficiently many times to exactly match the target data budget (i.e., $\text{subset size} \times \tau = \text{total data size}$). Unlike \cite{muennighoff2023scaling}, subsets are sampled independently for each run rather than sampling within the larger subsets, to mitigate sampling bias and balance stochasticity across conditions. To remain consistent with the Curriculum Learning setting of the main experiments, examples within each subset are ordered by increasing difficulty; across epochs, this difficulty schedule is preserved and repeated rather than reshuffled, as illustrated in Figure~\ref{fig:data_reuse_schema}. We also put the results for instruct model in Figure~\ref{fig:data_reuse_instruct}.
\section{Additional Experiment Results}
\begin{figure*}[ht]
    \centering
    % 佔位符: 請替換為 Base 模型的泛化結果圖 (多個子圖)
    \includegraphics[width=0.7\linewidth]{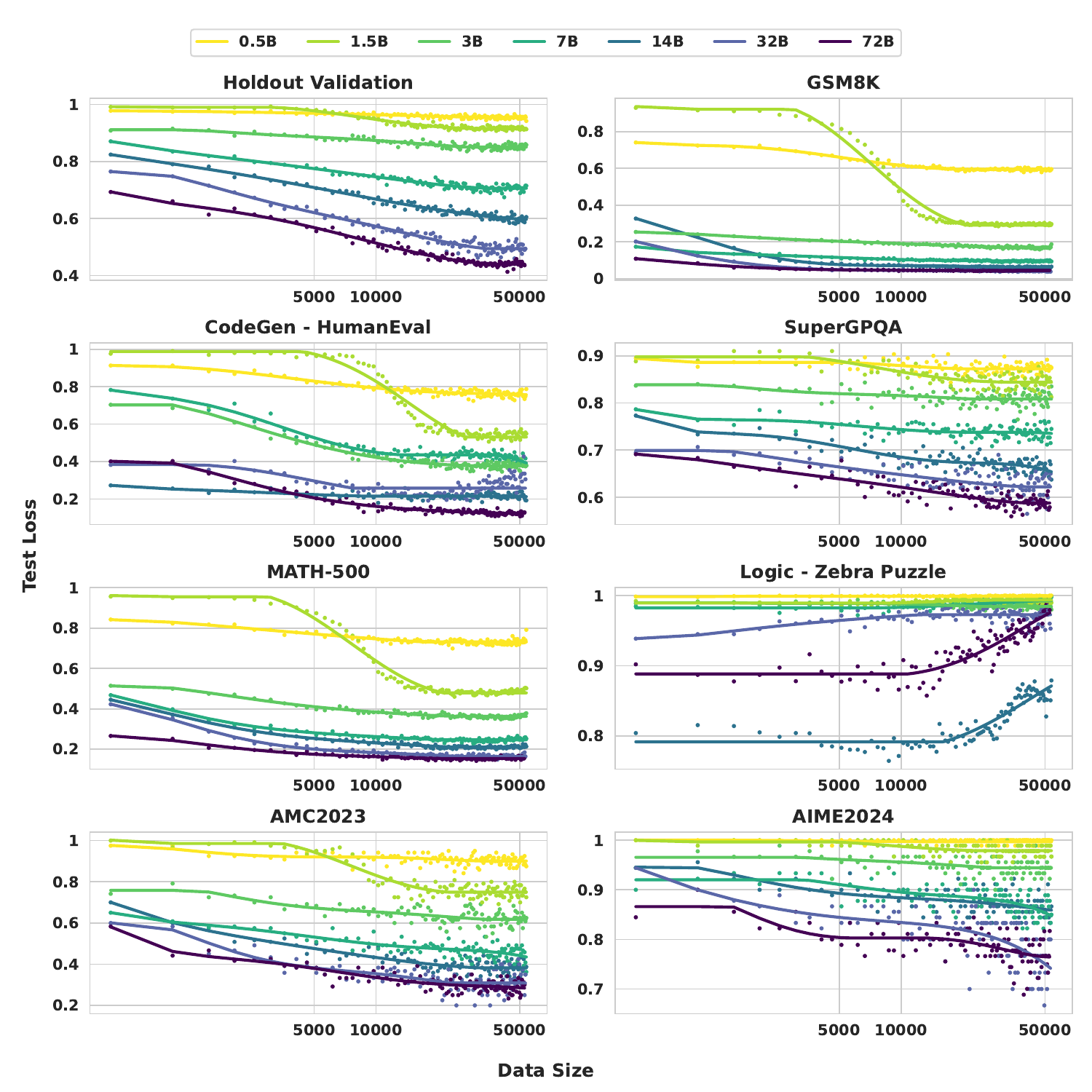}
    \caption{Test loss on in-domain and out-of-domain benchmarks vs data size for Base models. It shows modest positive transfer on in-domain tasks, with limited or negative transfer on OOD tasks.}
    \label{fig:generalization_results_base}
\end{figure*}
\label{app:additional_results}

This section provides supplementary experimental results that support and extend the analyses presented in the main body of the paper.

\subsection{Performance on In-Domain and Out-of-Domain Tasks}
\label{app:generalization_tasks}

\begin{figure*}[tb]
    \centering
    % 佔位符: 請替換為 Instruct 模型的泛化結果圖 (多個子圖)
    \includegraphics[width=0.7\linewidth]{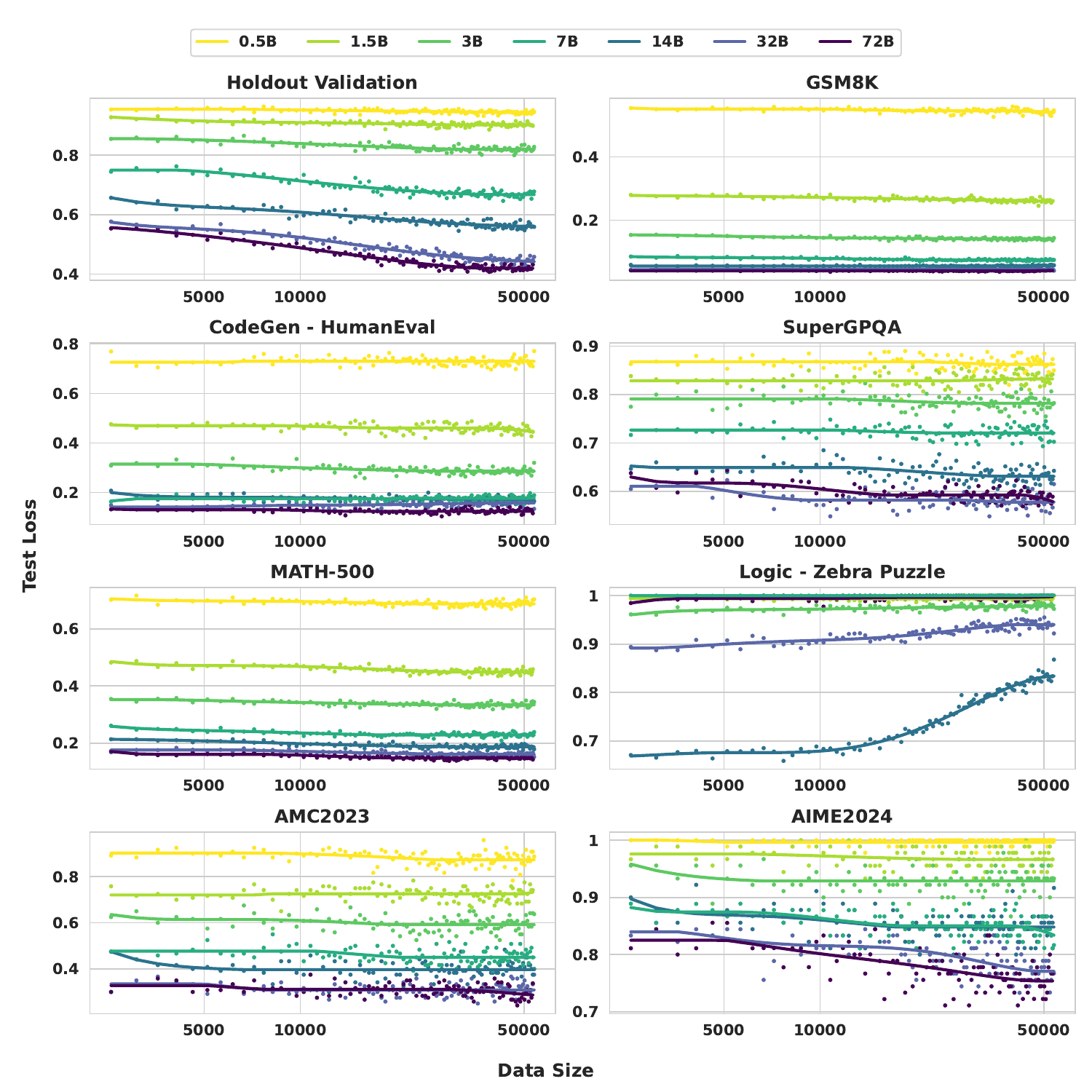}
    \caption{Test loss on in-domain and out-of-domain benchmarks vs data size for Instruct models.}
    \label{fig:generalization_results_instruct}
\end{figure*}

To assess how the mathematical reasoning capabilities acquired during RL fine-tuning generalize, we evaluated our models on a comprehensive suite of unseen benchmarks. We categorize these into two groups: in-domain different tasks (other mathematics datasets) and out-of-domain tasks (e.g., code, science, logic). The results are presented in Figure~\ref{fig:generalization_results_base} and Figure~\ref{fig:generalization_results_instruct}.

\textbf{In-Domain Generalization (Different Mathematical Tasks).}
On mathematics benchmarks not included in our training set (such as GSM8K, MATH, AIME, and AMC), we observe a generally positive transfer of learned skills. For most of these tasks, the test loss shows a modest but consistent decrease as training progresses, particularly for the larger models. This suggests that the model's enhanced reasoning ability is not overfitted to the training distribution and is applicable to a wider range of mathematical problems.

\textbf{Out-of-Domain Generalization.}
When evaluating on tasks outside of mathematics, the generalization is more limited. For both code generation (HumanEval) and science problems (SuperGPQA), performance remains largely static throughout the training process across all model sizes, with test loss curves staying flat. This indicates that the specialized mathematical reasoning skills do not readily transfer to these domains. A noteworthy phenomenon is observed in the logical reasoning task (Zebra Puzzle): the largest models (particularly the 14B variants) show a degradation in performance (an increase in test loss) as training progresses, suggesting a potential negative transfer effect where intensive optimization on mathematical reasoning may interfere with capabilities required for certain types of logical puzzles.
\subsection{Ablation on GRPO Hyperparameters}
\label{subsec:ablation_grpo}
% \begin{tcolorbox}[colback=blue!5!white, colframe=blue!30!white, title=Observation 6]
% A larger GRPO rollout group size ($G$) is consistently more data-efficient, while the compute-optimal $G$ grows with the total computational budget.
% \end{tcolorbox}
\begin{figure*}[t] % 注意这里改成了 figure*，位置建议用 [t] (top)
    \centering
    
    % --- 第一行 ---
    \begin{subfigure}[t]{0.48\textwidth} % 宽度调整为全页宽的 48%
        \centering
        \includegraphics[width=\linewidth]{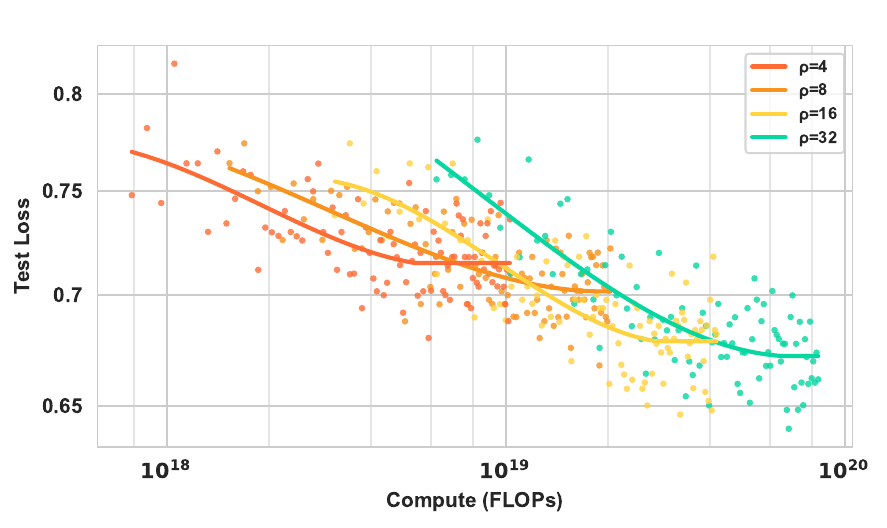}
        \caption{7B-Base: Loss vs.\ Compute}
        \label{fig:grpo_ablation_a}
    \end{subfigure}
    \hfill % 在两图之间填充空白，使其分别对齐左右边缘
    \begin{subfigure}[t]{0.48\textwidth}
        \centering
        \includegraphics[width=\linewidth]{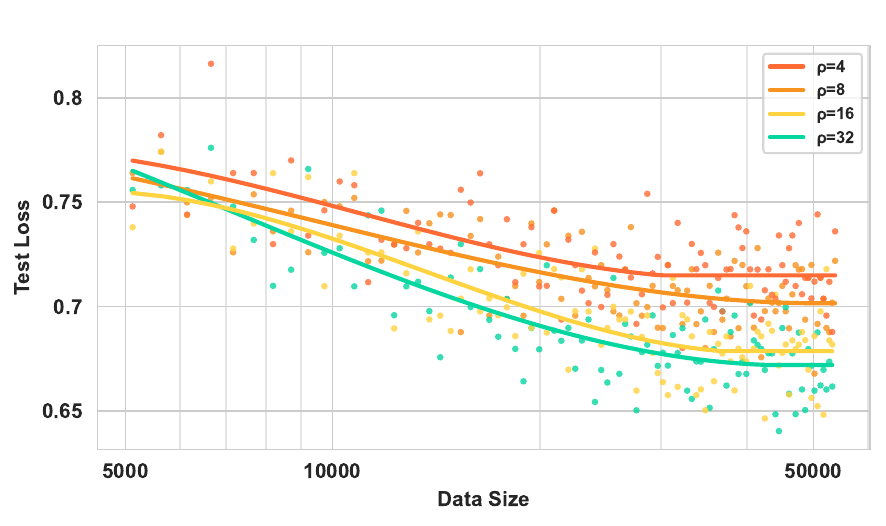}
        \caption{7B-Base: Loss vs.\ Data Size}
        \label{fig:grpo_ablation_b}
    \end{subfigure}
    
    \vspace{1em} % 增加第一行和第二行之间的垂直间距
    
    % --- 第二行 ---
    \begin{subfigure}[t]{0.48\textwidth}
        \centering
        \includegraphics[width=\linewidth]{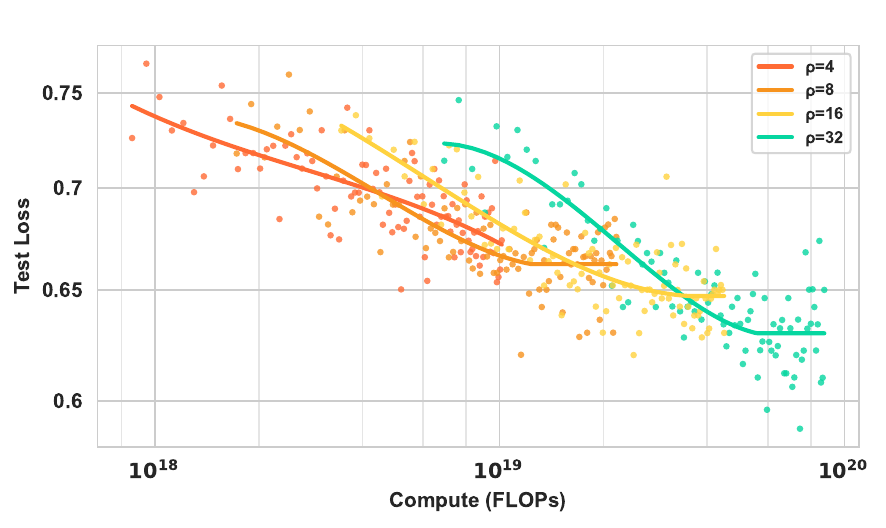}
        \caption{7B-Instruct: Loss vs.\ Compute}
        \label{fig:grpo_ablation_c}
    \end{subfigure}
    \hfill
    \begin{subfigure}[t]{0.48\textwidth}
        \centering
        \includegraphics[width=\linewidth]{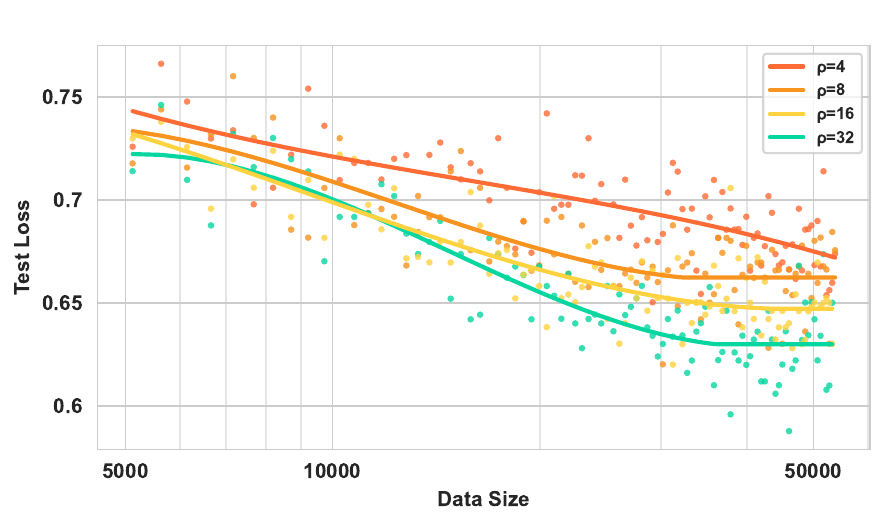}
        \caption{7B-Instruct: Loss vs.\ Data Size}
        \label{fig:grpo_ablation_d}
    \end{subfigure}

    \caption{
        Effects of GRPO rollout size on training efficiency.
        The top row shows the base model results, while the bottom row shows the instruct model results.
    }
    \label{fig:grpo_ablation}
\end{figure*}
We conducted an ablation study on the rollout group size $G$, a key GRPO hyperparameter that controls how many responses are sampled per prompt. This directly affects both the compute per update and the stability of the training signal. We tested $G \in \{4, 8, 16, 32\}$ on the 7B models.

\textbf{Data-centric View}. Figure~\ref{fig:grpo_ablation_b} and \ref{fig:grpo_ablation_d} shows that larger rollout sizes consistently yield better sample efficiency: $G=32$ achieves the lowest test loss for the same number of unique samples. This supports the intuition that more responses per question provide a stronger advantage estimate and thus more effective gradient updates.

\textbf{Compute-centric View}. The optimal rollout size $G$ is not fixed but shifts with the training budget. 
This implies that practitioners should tune $G$ according to available compute rather than relying on a universal setting. 
We attribute this dynamic to the trade-off between the higher variance reduction from larger $G$ and the additional FLOPs it consumes, which makes small $G$ preferable at low budgets but large $G$ superior when ample compute is available.
\subsection{Performance Compared with Advanced Models}
\label{app:sota_comparison}

\begin{figure}[t]
    \centering
    \includegraphics[width=0.9\linewidth]{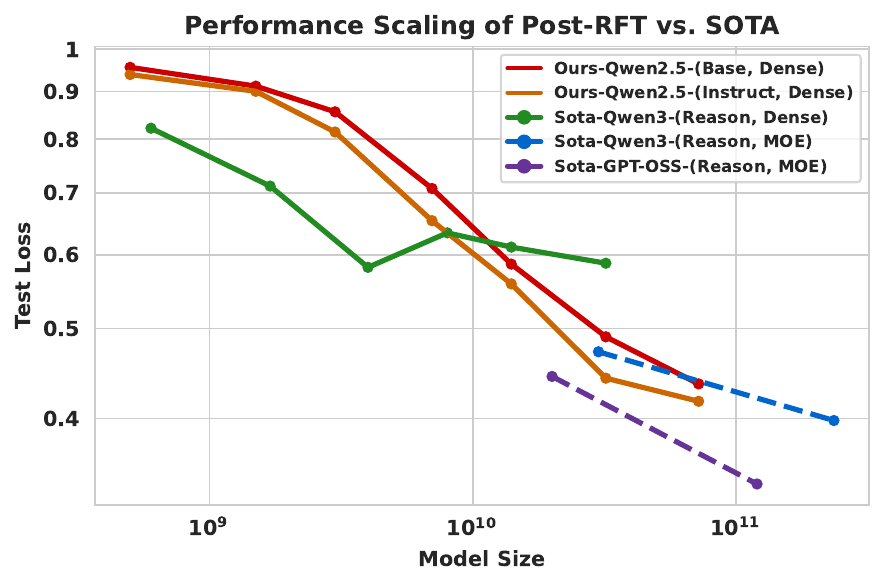}
    \caption{Relation between test loss and model size $N$ for our trained model and SOTA models.}
    \label{fig:loss_vs_model_size}
\end{figure}

We train models of varying sizes to convergence and compare their final test loss. As shown in Figure~\ref{fig:loss_vs_model_size}, larger models consistently achieve lower loss, improving monotonically with scale, suggesting diminishing returns at low parameter counts.

We also benchmark our RL-tuned Qwen2.5 models~\citep{qwen2025qwen25technicalreport} against state-of-the-art open-source reasoning systems, including Qwen3~\citep{yang2025qwen3technicalreport} and GPT-OSS~\citep{openai2025gptoss120bgptoss20bmodel}, detailed in Table~\ref{tab:sota_comparison_appendix}. On our held-out set, the 32B and 72B models match or surpass dense Qwen3 counterparts of similar size, highlighting the effectiveness of RL post-training. Mixture-of-experts models such as Qwen3 and GPT-OSS achieve approximate loss at much larger scales (235B), with GPT-OSS-120B currently leading.

\begin{table*}[!b]
\centering
\small
\setlength{\tabcolsep}{5pt}
\begin{tabular}{@{}l *{11}{c}@{}}
\toprule
& \multicolumn{11}{c}{\textbf{Pass@1 Score by Model Size}} \\
\cmidrule(lr){2-12}
\textbf{Model Family} & 0.5B$^{a}$ & 1.5B$^{a}$ & 3B$^{a}$ & 7B$^{a}$ & 14B & 20B & 30B$^{b}$ & 32B & 72B & 120B & 235B$^{c}$ \\
\midrule
Qwen2.5-Base     & 0.070 & 0.116 & 0.182 & 0.338 & 0.450 & --    & --    & 0.540 & 0.607 & --             & --             \\
Qwen2.5-Instruct & 0.078 & 0.138 & 0.216 & 0.380 & 0.488 & --    & --    & 0.590 & 0.617 & --             & --             \\
Qwen3            & 0.178 & 0.288 & 0.418 & 0.366 & 0.388 & --    & 0.528 & 0.412 & --    & --             & \textbf{0.602} \\
GPT-OSS          & --    & --    & --    & --    & --    & 0.556 & --    & --    & --    & \textbf{0.660} & --             \\
\bottomrule
\end{tabular}
\caption{Performance of various models on the held-out evaluation set (Pass@1). Our RL-tuned Qwen2.5 models are compared against external SOTA reasoning systems.
\textsuperscript{a}Qwen3 uses nearby sizes (0.6B, 1.7B, 4B, 8B) in these columns.
\textsuperscript{b}Qwen3-30B is the A3B MoE variant.
\textsuperscript{c}Qwen3-235B is the A22B MoE variant.}
\label{tab:sota_comparison_appendix}
\end{table*}

\section{Formula Fitting and Derivation}
\subsection{FLOPs Calculation Methodology}
\label{app:flops_calculation}
The computational cost for a LLM is primarily determined by the number of non-embedding parameters ($N$) and the number of processed tokens ($T$). The costs for the fundamental operations are:
\begin{itemize}
    \item \textbf{Forward Pass Cost:} The cost of a single forward pass is approximately $C_{\text{fwd}} \approx 2NT$ FLOPs.
    \item \textbf{Backward Pass Cost:} The backward pass is approximately twice as expensive as the forward pass, so $C_{\text{bwd}} \approx 4NT$ FLOPs.
\end{itemize}
A full training step, which includes one forward and one backward pass for the gradient update, therefore has a total computational cost of:
\begin{equation}
    C_{\text{train}} = C_{\text{fwd}} + C_{\text{bwd}} \approx 2NT + 4NT = 6NT \text{ FLOPs.}
\end{equation}
\begin{equation}
    \text{FLOPs}_{\text{step}} = 6 \times N \times T_{\text{step}}
\end{equation}
By recording the exact number of processed tokens $T$ per step, we compute the cumulative FLOPs reported throughout this paper as the sum of these per-step calculations over the course of training.

\subsection{Coefficient Comparison}
\label{app:coeff_comparison}
We consider the two laws
\begin{equation}
    \ln L(N, C) = -\,k_C(N)\,\ln C + E_{C}(N),
    \label{eq:compute_law_app}
\end{equation}
and
\begin{equation}
    \ln L(N, D) = -\,k_D(N)\,\ln D + E_{D}(N),
    \label{eq:data_law_app}
\end{equation}
are consistent under the linkage \(C = N D \phi\) where \(\phi>0\) is a constant for simplification.

\medskip
\noindent\textbf{Claim.}
Under \(C = N D \phi\), the slopes coincide and the intercepts differ by a known shift:
\begin{align}
    k_C(N) &= k_D(N) \;=\; k(N), \label{eq:k_equal}\\
    E_C(N) &= E_D(N) \;+\; k(N)\,\ln\!\bigl(N\phi\bigr). \label{eq:E_relation}
\end{align}

\medskip
\noindent\textit{Proof.}
Substitute \(C=N D \phi\) into \eqref{eq:compute_law_app}:
\begin{align*}
\ln L(N,C)
&= -k_C(N)\ln(N D \phi) + E_C(N) \\
&= -k_C(N)\bigl[\ln D + \ln(N\phi)\bigr] + E_C(N) \\
&= -k_C(N)\ln D \nonumber \\
&\quad + \Bigl(E_C(N) - k_C(N)\ln(N\phi)\Bigr).
\end{align*}
Comparing this with \eqref{eq:data_law_app}, i.e.,
\(\ln L(N,D) = -\,k_D(N)\,\ln D + E_D(N)\),
equality for all \(D>0\) forces the coefficients of \(\ln D\) and the constants to match:
\begin{align*}
k_D(N)=k_C(N)=:k(N),\\
E_D(N)=E_C(N)-k(N)\,\ln(N\phi).
\end{align*}
Rearranging the second identity yields \eqref{eq:E_relation}.

The observation from Figure~\ref{fig:coef_k_c} and Figure~\ref{fig:coef_k_d} also matches with this conclusion.
\subsection{Hyper-parameter Fitting}
We present the uncertainty analysis for raw data and fitting results for hyper parameters in Table~\ref{tab:uncertainty_comparison} and Table~\ref{tab:kn_parameters}.
\label{app:fitting_4_KE}
\begin{table*}[tb]
\centering
\begin{tabular}{llcccccc}
\hline
\multirow{2}{*}{\textbf{Family}} & \multirow{2}{*}{\textbf{Model}} & \multicolumn{3}{c}{\textbf{Base}} & \multicolumn{3}{c}{\textbf{Instruct}} \\
\cline{3-8}
& & Test Loss & Avg Std & SEM & Test Loss & Avg Std & SEM \\
\hline
\multirow{7}{*}{Qwen2.5} & 0.5B & 0.9419 & 0.0082 & 0.0048 & 0.9458 & 0.0073 & 0.0042 \\
& 1.5B & 0.9129 & 0.0091 & 0.0053 & 0.8988 & 0.0098 & 0.0057 \\
& 3B & 0.8582 & 0.0129 & 0.0074 & 0.8281 & 0.0112 & 0.0065 \\
& 7B & 0.7148 & 0.0147 & 0.0085 & 0.6777 & 0.0142 & 0.0082 \\
& 14B & 0.6051 & 0.0149 & 0.0086 & 0.5588 & 0.0143 & 0.0083 \\
& 32B & 0.4937 & 0.0056 & 0.0032 & 0.4579 & 0.0127 & 0.0073 \\
& 72B & 0.4359 & 0.0143 & 0.0082 & 0.4320 & 0.0140 & 0.0081 \\
\hline
\multirow{4}{*}{Llama~3} & 1B & -- & -- & -- & 0.9446 & 0.0075 & 0.0043 \\
& 3B & -- & -- & -- & 0.8842 & 0.0144 & 0.0083 \\
& 8B & -- & -- & -- & 0.7496 & 0.0257 & 0.0149 \\
& 70B & -- & -- & -- & 0.5499 & 0.0422 & 0.0244 \\
\hline
\end{tabular}
\caption{\tzl{Uncertainty Analysis for raw data: Qwen2.5 and Llama~3 Models (Holdout Score)}}
\label{tab:uncertainty_comparison}
\end{table*}

\begin{table}[tb]
\centering
\resizebox{\columnwidth}{!}{%
\begin{tabular}{llcccc}
\toprule
\textbf{Family} & \textbf{Configuration} & \textbf{Scenario} & $K_{max}$ & $N_0$ (B) & $R^2$ \\
\midrule
\multirow{8}{*}{Qwen2.5} & Base $L(N,C)$ & Intra & 0.135 & 13.1 & 0.996 \\
 & Base $L(N,C)$ & Inter & 0.152 & 17.4 & 0.994 \\
 & Base $L(N,D)$ & Intra & 0.135 & 11.5 & 0.995 \\
 & Base $L(N,D)$ & Inter & 0.163 & 17.0 & 0.995 \\
\cline{2-6}
 & Instruct $L(N,C)$ & Intra & 0.128 & 17.3 & 0.997 \\
 & Instruct $L(N,C)$ & Inter & 0.144 & 28.3 & 0.995 \\
 & Instruct $L(N,D)$ & Intra & 0.133 & 17.1 & 0.997 \\
 & Instruct $L(N,D)$ & Inter & 0.148 & 27.2 & 0.995 \\
\midrule
\multirow{4}{*}{Llama~3} & Instruct $L(N,C)$ & Intra & 0.089 & 11.3 & 0.998 \\
 & Instruct $L(N,C)$ & Inter & 0.074 & 8.5 & 0.995 \\
 & Instruct $L(N,D)$ & Intra & 0.091 & 12.7 & 0.998 \\
 & Instruct $L(N,D)$ & Inter & 0.087 & 11.8 & 0.997 \\
\bottomrule
\end{tabular}%
}
\caption{Comparison of $K_{max}$ and $N_0$ parameters across model families, fitting scenarios, and scaling dimensions. All fits achieve $R^2 > 0.99$.}
\label{tab:kn_parameters}
\end{table}
\section{A Loss Decomposition Model for Scaling Analysis}

During the analysis, we found a more generalized form of the potential scaling law function that fits the curves well.
This model fits the same dataset as the main experiments and is included here as a formally documented alternative for future research.

\subsection{Loss Decomposition Model}

\paragraph{The General Loss Decomposition.}
We construct the generalized formula as follows, based on the observation of the loss composition of post-training and the experiment data:

\begin{equation}
\label{eq:v3}
    L(N,D)
    = L_{\infty} 
    + G(N)
    + \lambda(N)
    \cdot P(N, D)
\end{equation}

Each term in Equation~\eqref{eq:v3} represents a clear part decomposing the loss:
\begin{itemize}[nosep,leftmargin=*]
    \item $L_{\infty}$ denotes the \textbf{irreducible loss}, the fundamental loss floor that persists even with infinite model capacity and unlimited data. It reflects task-intrinsic uncertainty and noise that cannot be eliminated by improved modeling or additional training.

    \item $G(N)$ denotes the \textbf{model-limited loss}, capturing the asymptotic loss floor imposed by finite model capacity $N$ in the limit of infinite data. It corresponds to the capacity-dependent performance frontier of models with size $N$.

    \item $\lambda(N)$ denotes the \textbf{learnable capacity}, the maximum achievable reduction in loss that a model of size $N$ can attain through post-training, beyond its model-limited loss. In RL post-training settings, this term depends on the pretraining regime and the degree of mismatch between pretraining and post-training task settings. The monotonic dependence of $\lambda(N)$ on model size is non-trivial, and its precise modeling likely requires additional assumptions and empirical characterization.

    \item $P(N,D)$ denotes the \textbf{learning progress}, a normalized function taking values in $[0,1]$ that quantifies the fraction of the learnable capacity $\lambda(N)$ realized when training on a dataset of size $D$.
\end{itemize}

\paragraph{Instantiated Model.}
Following prior empirical scaling law studies, we parameterize the model-limited loss as
\begin{equation}
    G(N) = \left(\frac{N_0}{N}\right)^{\alpha},
\end{equation}
reflecting the observed power-law dependence of loss on model size in the infinite-data regime~\citep{kaplan2020scaling}.

Empirically, learning curves exhibit an S-shaped transition when plotted in log–log coordinates (e.g. Figure~\ref{fig:intro_scaling_prediction}). 
Motivated by this observation, we model the learning progress term $P(N,D)$ as a logistic function in $\log D$,
\begin{equation}
    P(N,D) = \frac{1}{1 + \left(\frac{D}{D_0(N)}\right)^{\beta}},
\end{equation}
where $D_0(N)$ denotes the characteristic dataset scale at which half of the learnable capacity is realized, and is treated as a $N$-dependent parameter.

Combining the above components, we arrive at the following instantiated loss model:
\begin{equation}
\label{eq:v3_o2k}
    L(N,D)
    = L_{\infty} 
    + \left(\frac{N_0}{N}\right)^{\alpha}
    + \frac{\lambda(N)}{1 + \left(\frac{D}{D_0(N)}\right)^{\beta}}
\end{equation}

This is what we used to fit and extrapolation in Figure~\ref{fig:v3_pred}

\subsection{Predictability and Extrapolation}

\begin{figure*}[t]
    \centering
    \begin{subfigure}{0.48\textwidth}
        \centering
        \includegraphics[width=\linewidth]{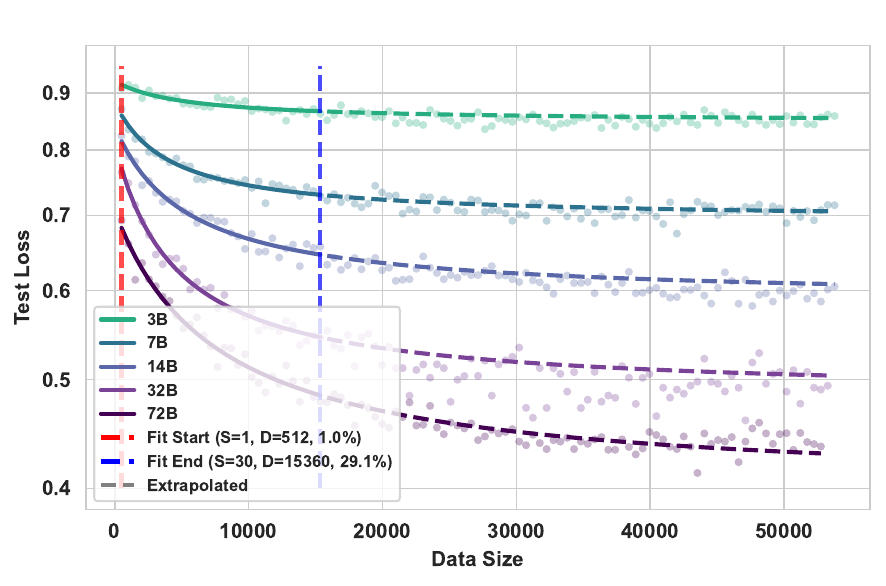}
        \caption{Intra-model prediction using the early 30\% of training steps.}
        \label{fig:v3_pred_a}
    \end{subfigure}
    \hfill
    \begin{subfigure}{0.48\textwidth}
        \centering
        \includegraphics[width=\linewidth]{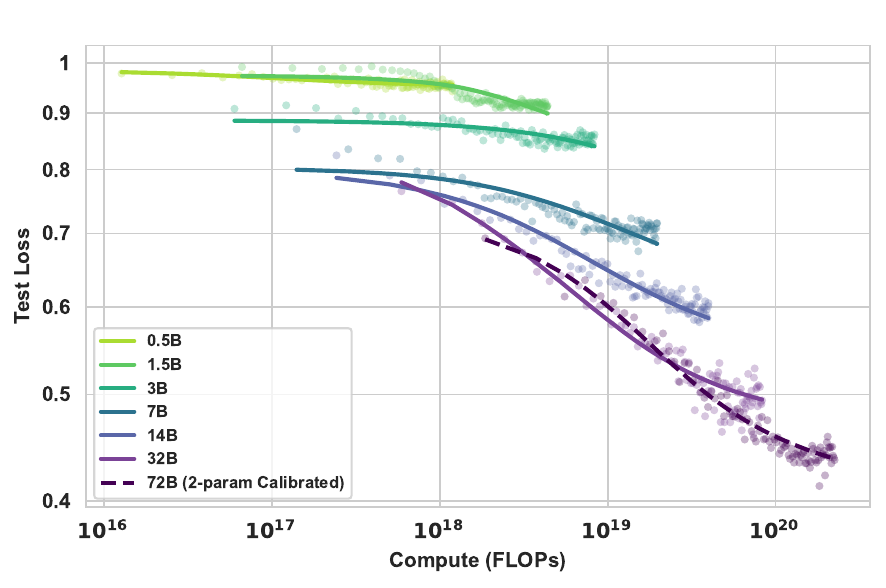}
        \caption{Inter-model extrapolation for 72B under fixed shared shape.}
        \label{fig:v3_pred_b}
    \end{subfigure}
    \caption{Extrapolation of the loss decomposition model (Equation~\eqref{eq:v3_o2k}).
    (a) Intra-model prediction using partial learning-curve observations.
    (b) Inter-model extrapolation for 72B with global exponents fixed from smaller models.}
    \label{fig:v3_pred}
\end{figure*}

We evaluate the extrapolation behavior of the loss–decomposition model (Equation~\eqref{eq:v3_o2k}) by applying it to the same experimental learning-curve data used in the main analysis, and by testing its performance under two complementary settings (Figure~\ref{fig:v3_pred}).

\paragraph{Intra-model prediction.}
Using only the first 30\% of training steps for each model, the fitted curves closely match the held-out portions(Figure~\ref{fig:v3_pred}a). This indicates that the internal S-shaped structure of the model provides a sufficiently strong inductive bias for completing a single learning curve from early observations. See Table~\ref{tab:fit_v3_intra} for detailed fitting result.

\paragraph{Inter-model extrapolation.}
We fit all global exponents and the model-limited term using models up to 32B, and then extrapolate the resulting shared functional form to 72B by calibrating only two model-specific parameters, $\lambda(72\text{B})$ and $D_0(72\text{B})$.
The extrapolated curve aligns well with the observed 72B trajectory across the full data range, reflecting that the functional shape inferred from smaller models remains compatible with larger-scale behavior under this light calibration.

The fit is reasonably strong, indicating that the proposed formulation may capture key structural tendencies of the underlying scaling behavior.

\begin{table}[H]
\centering
\caption{Fitting details for $L(N,D)$ using only 30\% of the training steps for each model.}
\label{tab:fit_v3_intra}
\begin{tabular}{lccc}
\toprule
\multirow{2}{*}{\textbf{Model Size}} & \multicolumn{3}{c}{\textbf{Base}} \\
\cmidrule(lr){2-4}
 & $D_0$ & $lambda$ & $R^2$ \\
\midrule
3B & 5109.6316 & 0.0734 & 0.995 \\
7B & 3725.6374 & 0.1882 & 0.995 \\
14B & 4554.7619 & 0.2520 & 0.995 \\
32B & 3576.0323 & 0.3279 & 0.995 \\
72B & 5861.6228 & 0.3084 & 0.995 \\
\bottomrule
\end{tabular}
\end{table}

% \begin{figure}
%     \centering
%     \includegraphics[width=\textwidth]{figures/fit_xh/fit_param_base_xh_N_E_D0_lambda_compare_source.pdf}
%     \caption{Fitted parameters}
%     \label{fig:placeholder}
% \end{figure}

\subsection{Discussion: Effective log--log slope}

To relate the loss decomposition model to the slope-based formulation used in the main text, we examine the local behavior of $L(N,D)$ in log--log coordinates with respect to the data scale $D$.  
Specifically, we define the effective slope
\[
k(N,D)\;:=\; -\,\frac{\partial \log L(N,D)}{\partial \log D},
\]
which corresponds to the exponent in a local power-law approximation of the form
$\log L \approx -k \log D + \text{const}$.

For the loss decomposition model (Equation~\eqref{eq:v3_o2k}), the induced $k(N,D)$ is a smooth function of $D$ that vanishes in both the low-data and high-data limits, and attains its maximum around the characteristic scale $D \approx D_0(N)$.  
Evaluating the slope at this point yields a natural definition of the maximal effective slope,

\begin{align}
\label{eq:v3_kn}
    k_{\max}(N)
    =
    \frac{K_{\max}}{1 + S(N)}, \\
    K_{\max}:= \frac{\beta}{2},\\
    S(N):=
    \frac{2\!\left(L_{\infty} + (N_0/N)^{\alpha}\right)}{\lambda(N)}.
\end{align}

The resulting $k_{\max}(N)$ depends only on the model size $N$ through the parameters of the loss decomposition, and is uniformly bounded by $K_{\max}=\beta/2$.

From this perspective, the slope function $k(N)$ adopted in the main analysis (Equation~\eqref{eq:efficiency_intro}) can be interpreted as a parsimonious, low-parameter approximation to the effective maximal slope $k_{\max}(N)$ in Eq.~\ref{eq:v3_kn}.
This establishes structural consistency between the two descriptions: the loss decomposition model potentially captures the finer-grained $(N,D)$-dependent behavior, while the main-text formulation summarizes its dominant $N$-dependent trend in a compact form.

\subsection{Conclusion}

The loss decomposition model captures key empirical characteristics of RL post-training scaling behavior.

However, several components remain underdetermined, including the construction of the learnable capacity $\lambda(N)$, the dependence of the characteristic dataset scale $D_0(N)$ on model size, the role of the pretraining process, and the impact of mismatches between pretraining and post-training task settings.

For these reasons, we present this model as an appendix-level discussion rather than a core component of the main text.
We encourage future work to build on this formulation to further investigate and refine scaling laws for LLM-based reinforcement learning.

% \section{Response Length}

% \begin{figure}[H]
%     \centering
%     % 如果是单栏排版，建议将宽度设为 \linewidth 或 0.9\linewidth
%     \includegraphics[width=\linewidth]{./figures/outputs/outputs/response_base_response_length_N_E_R.pdf}
%     \caption{\tzl{Response length vs.\ Data size (Base models).}}
%     \label{fig:limiting_plots_base}
% \end{figure}

% \begin{figure}[H]
%     \centering
%     \includegraphics[width=\linewidth]{./figures/outputs/outputs/response_instruct_response_length_N_E_R.pdf}
%     \caption{\tzl{Response length vs.\ Data size (Instruct models).}}
%     \label{fig:limiting_plots_instruct}
% \end{figure}

% \begin{figure}[H]
%     \centering
%     \includegraphics[width=0.48\textwidth]{./figures/response_length/base_vs_instruct_L-N-response_length_fitting_comparison_k.pdf}
%     % \includegraphics[width=0.48\textwidth]{./figures/response_length/base_vs_instruct_L-N-response_length_fitting_comparison_E0.pdf}
%     \caption{Fitted response length parameters vs. Model Size. Left: $k_\text{response}(N)$. Right: $E_\text{response}(N)$.}
%     \label{fig:limiting_plots_wholefitted}
% \end{figure}

% \begin{wrapfigure}{r}{0.45\textwidth}
%     \centering
%     \vspace{-1em} % 调整竖直位置，避免顶到段落
%     \includegraphics[width=0.95\linewidth]{./figures/datareuse/exp2_tau_rectangles.pdf}
%     \caption{Data Reuse Schema}
%     \label{fig:data_reuse_schematic}
% \end{wrapfigure}

% 1.  越大的模型response length span越大
% 2. 3B以上response 效率拐点
% 3. 3B以下response 效率反而略有降低，3B是可观测response效率最差model
% 4. 这里的k_response(N)趋势表示越大模型对于response的利用率越有效，对小模型来说增大response length对效果提升也没啥用，都是乱说
\section{Ethics Statement}
This work is foundational in nature, focusing on the scaling properties of large language models in the domain of mathematical reasoning. Our research exclusively utilizes publicly available and previously published resources, including open-source models (e.g., Qwen2.5) and established datasets (e.g.,guru-RL-92k), thereby mitigating concerns related to data privacy, human subjects, or the release of sensitive information. The application domain of mathematical problem-solving does not inherently present risks of direct societal harm. The primary ethical consideration associated with this work is the environmental impact of the computational resources required for large-scale model training, a challenge common to the field. We believe that by providing insights into efficient resource allocation, our work contributes positively to mitigating this concern for future research.
\section{The Use of Large Language Models}
We used Large Language Model (LLM) to refine our initial draft. This process included checking for obvious grammatical and syntactical errors, as well as making the language more formal and academic. We reviewed the content generated by the LLM to ensure that no prohibited generated content appeared in the article.
\end{document}